\documentclass[twoside]{article}
\pdfoutput=1 
\usepackage[accepted]{aistats2018}

\usepackage[utf8]{inputenc} % allow utf-8 input
\usepackage[T1]{fontenc}    % use 8-bit T1 fonts
\usepackage{hyperref}       % hyperlinks
\usepackage{url}            % simple URL typesetting
\usepackage{booktabs}       % professional-quality tables
\usepackage{amsfonts}       % blackboard math symbols
\usepackage{nicefrac}       % compact symbols for 1/2, etc.
\usepackage{microtype}      % microtypography

\usepackage[load-configurations=version-1]{siunitx} % newer version
\usepackage{graphicx}
\usepackage{subfig}
\usepackage{bm}
\usepackage[algo2e, ruled, vlined]{algorithm2e}
\usepackage{amsmath}
\usepackage{amsthm}
\usepackage{amssymb}

\usepackage{float}

\newtheorem{theorem}{Theorem}
\newtheorem{lemma}[theorem]{Lemma}

\newtheorem{prop}{Proposition}

\DeclareMathOperator{\Tr}{Tr}
\let\OLDthebibliography\thebibliography
\renewcommand\thebibliography[1]{
  \OLDthebibliography{#1}
  \setlength{\itemsep}{3pt}
}

\begin{document}

\twocolumn[

\aistatstitle{Scaling up the Automatic Statistician: Scalable Structure Discovery using Gaussian Processes}

\aistatsauthor{ Hyunjik Kim and Yee Whye Teh }

\aistatsaddress{University of Oxford, DeepMind \\ \texttt{\{hkim,y.w.teh\}@stats.ox.ac.uk}} ]

\begin{abstract}
Automating statistical modelling is a challenging problem in artificial intelligence. The Automatic Statistician takes a first step in this direction, by employing a kernel search algorithm with Gaussian Processes (GP) to provide interpretable statistical models for regression problems. However this does not scale due to its $O(N^3)$ running time for the model selection. We propose Scalable Kernel Composition (SKC), a scalable kernel search algorithm that extends the Automatic Statistician to bigger data sets. In doing so, we derive a cheap upper bound on the GP marginal likelihood that sandwiches the marginal likelihood with the variational lower bound . We show that the upper bound is significantly tighter than the lower bound and thus useful for model selection.
\end{abstract}

\section{Introduction}
\label{sec:intro}
Automated statistical modelling is an area of research in its early stages, yet it is becoming an increasingly important problem \cite{pmlr-v64-guyon_review_2016}. As a growing number of disciplines use statistical analyses and models to help achieve their goals, the demand for statisticians, machine learning researchers and data scientists is at an all time high. Automated systems for statistical modelling serves to assist such human resources, if not as a best alternative where there is a shortage.

An example of a fruitful attempt at automated statistical modelling in nonparametric regression is Compositional Kernel Search (CKS) \cite{duvenaud2013structure}, an algorithm that fits a Gaussian Process (GP) to the data and automatically chooses a suitable parametric form of the kernel. This leads to high predictive performance that matches kernels hand-selected by GP experts \cite{rasmussen2006gaussian}. There also exist other approaches that tackle this model selection problem by using a more flexible kernel \cite{bach2004multiple,oliva2016bayesian,samo2015generalized, wilson2013gaussian,wilson2016deep}. However the distinctive feature of Duvenaud et al \cite{duvenaud2013structure} is that the resulting models are interpretable; the kernels are constructed in such a way that we can use them to describe patterns in the data, and thus can be used for automated exploratory data analysis. Lloyd et al \cite{lloyd2014automatic} exploit this to generate natural language analyses from these kernels, a procedure that they name Automatic Bayesian Covariance Discovery (ABCD). The Automatic Statistician\footnote{See \url{http://www.automaticstatistician.com/index/} for example analyses.} implements this to output a 10-15 page report when given data input. 

However, a limitation of ABCD is that it does not scale; due to the $O(N^3)$ time for inference in GPs, the analysis is constrained to small data sets, specialising in one dimensional time series data. This is undesirable not only because the average size of data sets is growing fast, but also because there is potentially more information in bigger data, implying a greater need for more expressive models that can discover finer structure. This paper proposes Scalable Kernel Composition (SKC), a scalable extension of CKS, to push the boundaries of automated interpretable statistical modelling to bigger data. In summary, our work makes the following contributions:
\begin{itemize}
    \item We propose the first scalable version of the Automatic Statistician that scales up to medium-sized data sets by reducing algorithmic complexity from $O(N^3)$ to $O(N^2)$ and enhancing parallelisability. 
    \item We derive a novel cheap upper bound on the GP marginal likelihood, that is used in SKC with the variational lower bound \cite{titsias2009variational} to sandwich the GP marginal likelihood.
    \item We show that our upper bound is significantly tighter than the lower bound, and plays an important role for model selection.
\end{itemize}

\section{ABCD and CKS}
\label{sec:abcd}
The Compositional Kernel Search (CKS) algorithm \cite{duvenaud2013structure} builds on the idea that the sum and product of two positive definite kernels are also positive definite. Starting off with a set $\mathcal{B}$ of base kernels defined on $\mathbb{R}\times\mathbb{R}$, the algorithm searches through the space of zero-mean GPs with kernels that can be expressed in terms of sums and products of these base kernels. $\mathcal{B}=\{\text{SE,LIN,PER}\}$ is used, which correspond to the squared exponential, linear and periodic kernel respectively (see Appendix \ref{apd:base} for the exact form of these base kernels). Thus candidate kernels form an open-ended space of GP models, allowing for an expressive model. Such approaches for structure discovery have also appeared in \cite{gardner17a, grosse2012exploiting}. A greedy search is employed to explore this space, with each kernel scored by the Bayesian Information Criterion (BIC) \cite{schwarz1978estimating} \footnote{BIC = log marginal likelihood with a model complexity penalty. We use a definition where higher BIC means better model fit. See Appendix \ref{apd:bic}.} after optimising the kernel hyperparameters by type II maximum likelihood (ML-II). See Appendix \ref{apd:kernelsearch} for the algorithm in detail.

The resulting kernel can be simplified to be expressed as a sum of products of base kernels, which has the notable benefit of interpretability. In particular, note $f_1 \sim GP(0,k_1), f_2 \sim GP(0,k_2) \Rightarrow f_1+f_2 \sim GP(0,k_1+k_2)$ for independent $f_1$ and $f_2$. So a GP whose kernel is a sum of products of kernels can be interpreted as sums of GPs each with structure given by the product of kernels. Now each base kernel in a product modifies the model in a consistent way. For example, multiplication by SE converts global structure into local structure since SE($x,x'$) decreases exponentially with $|x-x'|$, and multiplication by PER is equivalent to multiplication of the modeled function by a periodic function (see Lloyd et al \cite{lloyd2014automatic} for detailed interpretations of different combinations). This observation is used in Automatic Bayesian Covariance Discovery (ABCD) \cite{lloyd2014automatic}, giving a natural language description of the resulting function modeled by the composite kernel. In summary ABCD consists of two algorithms: the compositional kernel search CKS, and the natural language translation of the kernel into a piece of exploratory data analysis.

\section{Scaling up ABCD}
\label{sec:scaling}

ABCD provides a framework for a natural extension to big data settings, in that we only need to be able to scale up CKS, then the natural language description of models can be directly applied. The difficulty of this extension lies in the $O(N^3)$ time for evaluation of the GP marginal likelihood and its gradients with respect to the kernel hyperparameters. 

A na{\"i}ve approach is to subsample the data to reduce $N$, but then we may fail to capture global structure such as periodicities with long periods or omit a set of points displaying a certain local structure. We show failure cases of random subsampling in Section \ref{sec:largeexp}. Regarding more strategic subsampling, the possibility of a generic subsampling algorithm for GPs that is able to capture the aforementioned properties of the data is a challenging research problem in itself.

Alternatively it is tempting to use either an approximate marginal likelihood or the exact marginal likelihood of an approximate model as a proxy for the exact likelihood \cite{quinonero2005unifying,seeger2003fast,snelson2005sparse,titsias2009variational}, especially with modern GP approximations scaling to large data sets with millions of data points \cite{hensman2013gaussian, wilson2015msgp}. However such scalable GP methods are limited in interpretability as they often behave very differently to the full GP, lacking guarantees for the chosen kernel to faithfully reflect the actual structure in the data. In other words, the real challenge is to scale up the GP while preserving interpretability, and this is a difficult problem due to the tradeoff between scalability and accuracy of GP approximations. Our work pushes the frontiers of interpretable GPs to medium-sized data ($N=10K \sim 100K$) by reducing the computational complexity of ABCD from $O(N^3)$ to $O(N^2)$. Extending this to large data sets ($N=100K \sim 1M$) is a difficult open problem.

Our approach is as follows: we provide a cheap lower bound and upper bound to sandwich the exact marginal likelihood, and we use this interval for model selection. To do so we give a brief overview of the relevant work on low rank kernel approximations used for scaling up GPs, and we later outline how they can be applied to obtain cheap lower and upper bounds. 

\subsection{Nystr{\"o}m Methods and Sparse GPs}
The Nystr{\"o}m Method \cite{drineas2005nystrom, williams2001using} selects a set of $m$ inducing points in the input space $\mathbb{R}^D$ that attempt to explain all the covariance in the Gram matrix of the kernel; the kernel is evaluated for each pair of inducing points and also between the inducing points and the data, giving matrices $K_{mm},K_{mN}=K_{Nm}^\top$. This is used to create the Nystr{\"o}m approximation $\hat{K}=K_{Nm}(K_{mm})^{\dagger}K_{mN}$ of $K=K_{NN}$, where $\dagger$ is the pseudo-inverse. Applying Cholesky decomposition to $K_{mm}$, we see that the approximation admits the low-rank form $\Phi^\top\Phi$ and so allows efficient computation of determinants and inverses in $O(Nm^2)$ time (see Appendix \ref{apd:matrixid}). We later use the Nystr{\"o}m approximation to give an upper bound on the exact log marginal likelihood.
%However the approximation quality depends heavily on the choice of inducing points. Traditionally they are chosen to be a subset of the data. 

The Nystr{\"o}m approximation arises naturally in the sparse GP literature, where certain distributions are approximated by simpler ones involving $\boldsymbol{f_m}$, the GP evaluated at the $m$ inducing points: the Deterministic Training Conditional (DTC) approximation \cite{seeger2003fast} defines a model that gives the marginal likelihood $q(\mathbf{y})=\mathcal{N}(\mathbf{y}|0,\hat{K}+\sigma^2 I)$ ($\mathbf{y}$ is the vector of observations), whereas the Fully Independent Conditional (FIC) approximation \cite{snelson2005sparse} gives $q(\mathbf{y})=\mathcal{N}(\mathbf{y}|0,\hat{K}+\text{diag}(K-\hat{K})+\sigma^2 I)$, correcting the Nystr{\"o}m approximation along the diagonals. The Partially Independent Conditional (PIC) approximation \cite{quinonero2005unifying} further improves this by correcting the Nystr{\"o}m approximation on block diagonals, with blocks typically of size $m \times m$. Note that the approximation is no longer low rank for FIC and PIC, but matrix inversion can still be computed in $O(Nm^2)$ time by Woodbury's Lemma (see Appendix \ref{apd:matrixid}). 

The variational inducing points method (VAR) \cite{titsias2009variational} is rather different to DTC/FIC/PIC in that it gives the following variational lower bound on the exact log marginal likelihood:
\vspace*{-2mm}
\begin{equation} \label{eq:varlb}
\log[\mathcal{N}(\mathbf{y}|0,\hat{K}+\sigma^2 I)]-\frac{1}{2\sigma^2}\Tr(K-\hat{K})
\vspace*{-1mm}
\end{equation}
This lower bound is optimised with respect to the inducing points and the kernel hyperparameters, which is shown in the paper to successfully yield tight lower bounds in $O(Nm^2)$ time for reasonable values of $m$. Another useful property of VAR is that the lower bound can only increase as the set of inducing points grows \cite{matthews2015sparse,titsias2009variational}. It is also known that VAR always improves with extra computation, and that it successfully recovers the true posterior GP in most cases, contrary to other sparse GP methods \cite{bauer2016understanding}. Hence this is what we use in SKC to obtain a lower bound on the marginal likelihood and optimise the hyperparameters. We use 10 random initialisations of hyperparameters and choose the one with highest lower bound after optimisation. 
% Also note that contrary to DTC/FIC/PIC, (\ref{eq:varlb}) cannot be seen as the log marginal likelihood with a plug-in estimate for the Gram matrix.
% Approximating $p(\boldsymbol{f}|\boldsymbol{f_m},y)$ by $p(\boldsymbol{f}|\boldsymbol{f_m})$ and using a free Gaussian approximation $\phi(\boldsymbol{f_m})$ to $p(\boldsymbol{f_m}|y)$, we obtain the following variational lower bound 

\subsection{A cheap and tight upper bound on the log marginal likelihood}
Fixing the hyperparameters to be those tuned by VAR, we seek a cheap upper bound to the exact marginal likelihood. Upper bounds and lower bounds are qualitatively different, and in general it is more difficult to obtain an upper bound than a lower bound for the following reason: first note that the marginal likelihood is the integral of the likelihood with respect to the prior density of parameters. Hence to obtain a lower bound it suffices to exhibit regions in the parameter space giving high likelihood. However, to obtain an upper bound one must demonstrate the absence or lack of likelihood mass outside a certain region, an arguably more difficult task. There has been some work on the subject \cite{beal2003variational,ji2010bounded}, but to the best of our knowledge there has not been any work on cheap upper bounds to the marginal likelihood in large $N$ settings. So finding an upper bound from the perspective of the marginal likelihood can be difficult. Instead, we exploit the fact that the GP marginal likelihood has an analytic form, and treat it as a function of $K$. The GP log marginal likelihood is composed of two terms and a constant:
\vspace*{-1mm}
\begin{align}
\log p(y) =& \log[\mathcal{N}(y|0,K+\sigma^2 I)] \nonumber \\
=& -\frac{1}{2}\log \det (K+\sigma^2 I)  \nonumber \\
& -\frac{1}{2}y^\top(K+\sigma^2 I)^{-1}y-\frac{N}{2}\log(2\pi)
\end{align}
We give separate upper bounds on the negative log determinant (NLD) term and the negative inner product (NIP) term. For NLD, it has been proven that
\vspace*{-1mm}
\begin{equation}
    -\frac{1}{2}\log \det (K+\sigma^2 I) \leq -\frac{1}{2}\log \det (\hat{K} +\sigma^2 I)
\vspace*{-1mm}    
\end{equation}
a consequence of $K-\hat{K}$ being a Schur complement of $K$ and hence positive semi-definite (e.g. \cite{bardenet2015infdpp}). So the Nystr{\"o}m approximation plugged into the NLD term serves as an upper bound that can be computed in $O(Nm^2)$ time (see Appendix \ref{apd:matrixid}).

As for NIP, we point out that $\lambda y^\top(K+\sigma^2 I)^{-1}y = \min_{f \in \mathcal{H}} \sum_{i=1}^N (y_i-f(x_i))^2 + \lambda\|f\|_{\mathcal{H}}^2$, the optimal value of the objective function in kernel ridge regression where $\mathcal{H}$ is the Reproducing Kernel Hilbert Space associated with $k$ (e.g. \cite{murphy2012machine}). The dual problem, whose objective function has the same optimal value, is $\max_{\alpha \in \mathbb{R}^N} -\lambda[\alpha^\top(K+\sigma^2 I)\alpha-2\alpha^\top y]$. So we have the following upper bound:
\begin{equation} \label{eq:cg}
-\frac{1}{2}y^\top(K+\sigma^2 I)^{-1}y \leq \frac{1}{2}\alpha^\top(K+\sigma^2 I)\alpha-\alpha^\top y 
\end{equation}
$\forall \alpha \in \mathbb{R}^N$. Note that this is also in the form of an objective for conjugate gradients (CG) \cite{shewchuk1994introduction}, hence equality is obtained at the optimal value $\hat{\alpha}=(K+\sigma^2 I)^{-1}y$. We can approach the optimum for a tighter bound by using CG or preconditioned CG (PCG) for a fixed number of iterations to get a reasonable approximation to $\hat{\alpha}$. Each iteration of CG and the computation of the upper bound takes $O(N^2)$ time, but PCG is very fast even for large data sets and using FIC/PIC as the preconditioner gives fastest convergence in general \cite{cutajar_preconditioning_2016}. 

Recall that although the lower bound takes $O(Nm^2)$ to compute, we need $m=O(N^{\beta})$ for accurate approximations, where $\beta$ depends on the data distribution and kernel \cite{rudi2015less}. $\beta$ is usually close to 0.5, hence the lower bound is also effectively $O(N^2)$. In practice, the upper bound evaluation seems to be a little more expensive than the lower bound evaluation, but we only need to compute the upper bound once, whereas we must evaluate the lower bound and its gradients multiple times for the hyperparameter optimisation. We later confirm in Section \ref{sec:lbubexp} that the upper bound is fast to compute relative to the lower bound optimisation. We also show empirically that the upper bound is tighter than the lower bound in Section \ref{sec:lbubexp}, and give the following sufficient condition for this to be true:
\begin{prop}
Suppose $(\hat{\lambda}_i)_{i=1}^N$ are the eigenvalues of $\hat{K}+\sigma^2 I$ in descending order. Then if (P)CG for the NIP term converges and $\hat{\lambda}_N \geq 2\sigma^2$, then the upper bound is tighter than the lower bound.
\end{prop}
Notice that $\hat{\lambda}_N \geq \sigma^2 \hspace{2mm} \forall \hat{K},$ so the assumption is feasible. The proof is in Appendix \ref{apd:proof}. 

We later show in Section \ref{sec:exp} that the upper bound is not only tighter than the lower bound, but also much less sensitive to the choice of inducing points. Hence we use the upper bound to choose between kernels whose BIC intervals overlap.

\begin{algorithm2e}[t]
\caption{Scalable Kernel Composition (SKC)}
\label{alg:skc}
\KwIn{data $x_1, \ldots, x_n \in \mathbb{R}^D, y_1, \ldots, y_n \in \mathbb{R}$, base kernel set $\mathcal{B}$, depth $d$, number of inducing points $m$, kernel buffer size $S$.}
\KwOut{$k$, the resulting kernel.}
For each base kernel on each dimension, obtain lower and upper bounds to BIC (BIC interval), set $k$ to be the kernel with highest upper bound, and add $k$ to kernel buffer $\mathcal{K}$. \\
$\mathcal{C} \leftarrow \emptyset$ \\
\For{depth=1:d}{
  From $\mathcal{C}$, add to $\mathcal{K}$ all kernels whose intervals overlap with $k$ if there are fewer than $S$ of them, else add the kernels with top $S$ upper bounds. \\
  \For{$k' \in \mathcal{K}$}{
  Add following kernels to $\mathcal{C}$ and obtain their BIC intervals: \\
  \hspace{5mm}(1) All kernels of form $k'+B$ where $B$ is any base kernel on any dimension \\
  \hspace{5mm}(2) All kernels of form $k'\times B$ where $B$ is any base kernel on any dimension \\
  }
  \If{$ \exists k^* \in \mathcal{C}$ with higher upper bound than $k$}{
  $k \leftarrow k^*$
  }
}
\vspace*{-2mm}
\end{algorithm2e}

\subsection{SKC: Scalable Kernel Composition using the lower and upper bound} \label{subsec:skc}

We base our algorithm on two intuitive claims. First, the lower and upper bounds converge to the true log marginal likelihood for fixed hyperparameters as the number of inducing points $m$ increases. Second, the hyperparameters optimised by VAR converge to those obtained by optimising the exact log marginal likelihood as $m$ increases. The former is confirmed in Figure \ref{fig:vary_indpts} as well as in other works (e.g. \cite{bauer2016understanding} for the lower bound), and the latter is confirmed in Appendix \ref{apd:hyp_conv}.

The algorithm proceeds as follows: for each base kernel and a fixed value of $m$, we compute the lower and upper bounds to obtain an interval for the GP marginal likelihood and hence the BIC of the kernel, with its hyperparameters optimised by VAR. We rank these kernels by their intervals, using the upper bound as a tie-breaker for kernels with overlapping intervals. We then perform a semi-greedy kernel search, expanding the search tree on all (or some, controlled by buffer size $S$) kernels whose intervals overlap with the top kernel at the current depth. We recurse to the next depth by computing intervals for these child kernels (parent kernel +/$\times$ base kernel), ranking them and further expanding the search tree. This is summarised in Algorithm \ref{alg:skc}, and Figure \ref{fig:solar_tree} in Appendix \ref{apd:plots} is a visualisation of the tree for different values of $m$. Details on the optimisation and initialisation are given in Appendices \ref{apd:opt} and \ref{apd:hyp_init_priors}.

The hyperparameters found by VAR may not be the global maximisers of the exact GP marginal likelihood, but as in ABCD we can optimise the marginal likelihood with multiple random seeds and choose the local optimum closest to the global optimum. One may still question whether the hyperparameter values found by VAR agree with the structure in the data, such as period values and slopes of linear trends. We show in Sections \ref{sec:smallexp} and \ref{sec:largeexp} that a small $m$ suffices for this to be the case.

\textbf{Choice of m} We can guarantee that the lower bound increases with larger $m$, but cannot guarantee that the upper bound decreases, since we tend to get better hyperparameters for higher $m$ that boost the marginal likelihood and hence the upper bound. We verify this in Section \ref{sec:lbubexp}. Hence throughout SKC we fix $m$ to be the largest possible value that one can afford, so that the marginal likelihood with hyperparameters optimised by VAR is as close as possible to the marginal likelihood with optimal hyperparameters. It is natural to wonder whether an adaptive choice of $m$ is possible, using higher $m$ for more promising kernels to tighten their bounds. However a fair comparison of different kernels via this lower and upper bound requires that they have the same value of $m$, since using a higher $m$ is guaranteed to give a higher lower bound.
%Alternatively, we also tried building up $m$ from smaller values by optimising hyperparameters and each $m$ and using these as initialisers for the next $m$, but this gave similar results to just using the biggest possible $m$. So we use the latter that is more computationally efficient.

\begin{figure*}[h!]
  \centering
  \subfloat[Solar: fix inducing points]{\includegraphics[width=0.5\textwidth]{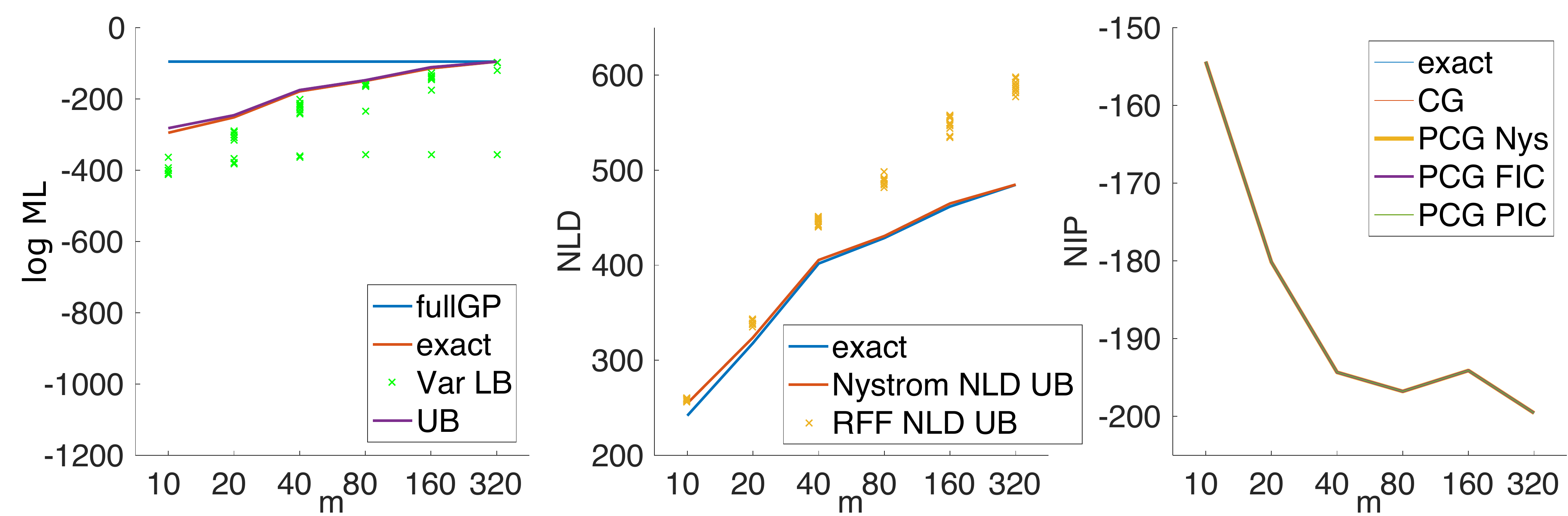}\label{fig:exp_solar_subset_half}}
  \hfill
  \subfloat[Solar: learn inducing points]{\includegraphics[width=0.5\textwidth]{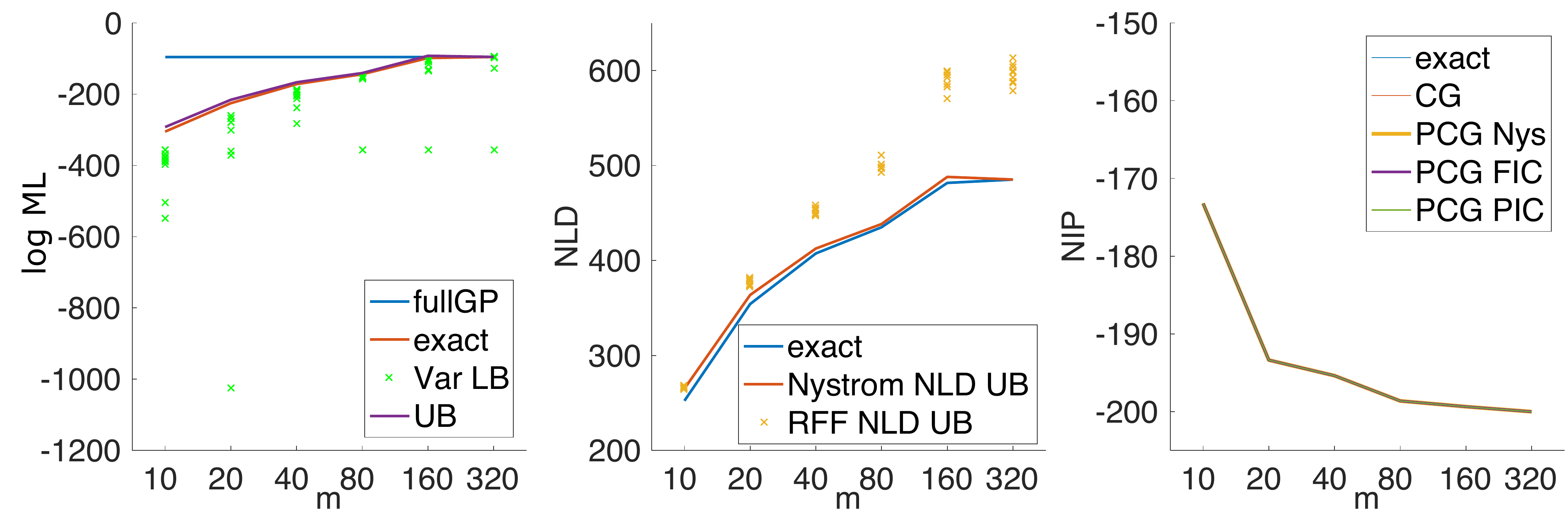}\label{fig:exp_solar_learn_ind_pts}}
  \caption{(a) Left: log marginal likelihood (ML) for fullGP with optimised hyperparameters, optimised VAR LB for each of 10 random initialisations per $m$, exact log ML for best hyperparameters out of 10, and corresponding UB. Middle: exact NLD and UB. Right: exact NIP and UB after $m$ iterations of CG/PCG. (b) Same as Figure \ref{fig:exp_solar_subset_half}, except learning inducing points for the LB optimisation and using them for subsequent computations.}
\end{figure*}

\textbf{Parallelisability} Note that SKC is extremely parallelisable across different random initialisations and different kernels at each depth, as is CKS. In fact the presence of intervals for SKC and hence buffers of kernels allows further parallelisation over kernels of different depths in certain cases (see Appendix \ref{apd:parallel}). 

\section{Experiments}
\label{sec:exp}
\subsection{Investigating the behaviour of the lower bound (LB) and upper bound (UB)} \label{sec:lbubexp}

We present results for experiments showing the bounds we obtain for two small time series and a multidimensional regression data set, for which CKS is feasible. The first is the annual solar irradiance data from 1610 to 2011, with 402 observations \cite{lean1995reconstruction}. The second is the time series Mauna Loa CO2 data \cite{mauna} with 689 observations. See Appendix \ref{apd:mauna_solar_plots_hyps} for plots of the time series. The multidimensional data set is the concrete compressive strength data set with 1030 observations and 8 covariates \cite{yeh1998modeling}. The functional form of kernels used for each of these data sets have been found by CKS (see Figure \ref{fig:cks_skd_comparison}). All observations and covariates have been normalised to have mean 0 and variance 1.

From the left of Figures \ref{fig:exp_solar_subset_half}, \ref{fig:exp_mauna_subset_half} and \ref{fig:exp_concrete_subset_half}, (the latter two can be found in Appendix \ref{apd:plots}) we see that VAR gives a LB for the optimal log marginal likelihood that improves with increasing $m$. The best LB is tight relative to the exact log marginal likelihoods at the hyperparameters optimised by VAR. We also see that the UB is even tighter than the LB, and increases with $m$ as hypothesised. From the middle plots, we observe that the Nystr{\"o}m approximation gives a very tight UB on the NLD term. We also tried using RFF to get a stochastic UB  (see Appendix \ref{apd:rff_ub}), but this is not as tight, especially for larger values of $m$. From the right plots, we can see that PCG with any of the three preconditioners (Nystr{\"o}m, FIC, PIC) give a very tight UB to the NIP term, whereas CG may require more iterations to get tight, for example in Figures \ref{fig:exp_mauna_subset_half}, \ref{fig:exp_mauna_learn_ind_pts} in Appendix \ref{apd:plots}.

Figure \ref{fig:vary_indpts} shows further experimental evidence for a wider range of kernels reinforcing the claim that the UB is tighter than the LB, as well as being much less sensitive to the choice of inducing points. This is why the UB is a much more reliable metric for the model selection than the LB. Hence we use the UB for a one-off comparison of kernels when their BIC intervals overlap. 

Comparing Figures \ref{fig:exp_solar_subset_half}, \ref{fig:exp_mauna_subset_half}, \ref{fig:exp_concrete_subset_half} against \ref{fig:exp_solar_learn_ind_pts}, \ref{fig:exp_mauna_learn_ind_pts}, \ref{fig:exp_concrete_learn_ind_pts}, learning inducing points does not lead to a vast improvement in the VAR LB. In fact the differences are not very significant, and sometimes learning inducing points can get the LB stuck in a bad local minimum, as indicated by the high variance of LB in the latter three figures. Moreover the differences in computational time is significant as we can see in Table \ref{tab:times} of Appendix \ref{apd:times}. Hence the computation-accuracy trade-off is best when fixing the inducing points to a random subset of training data. 

\begin{figure*}
  \centering
  \includegraphics[width=0.9\linewidth]{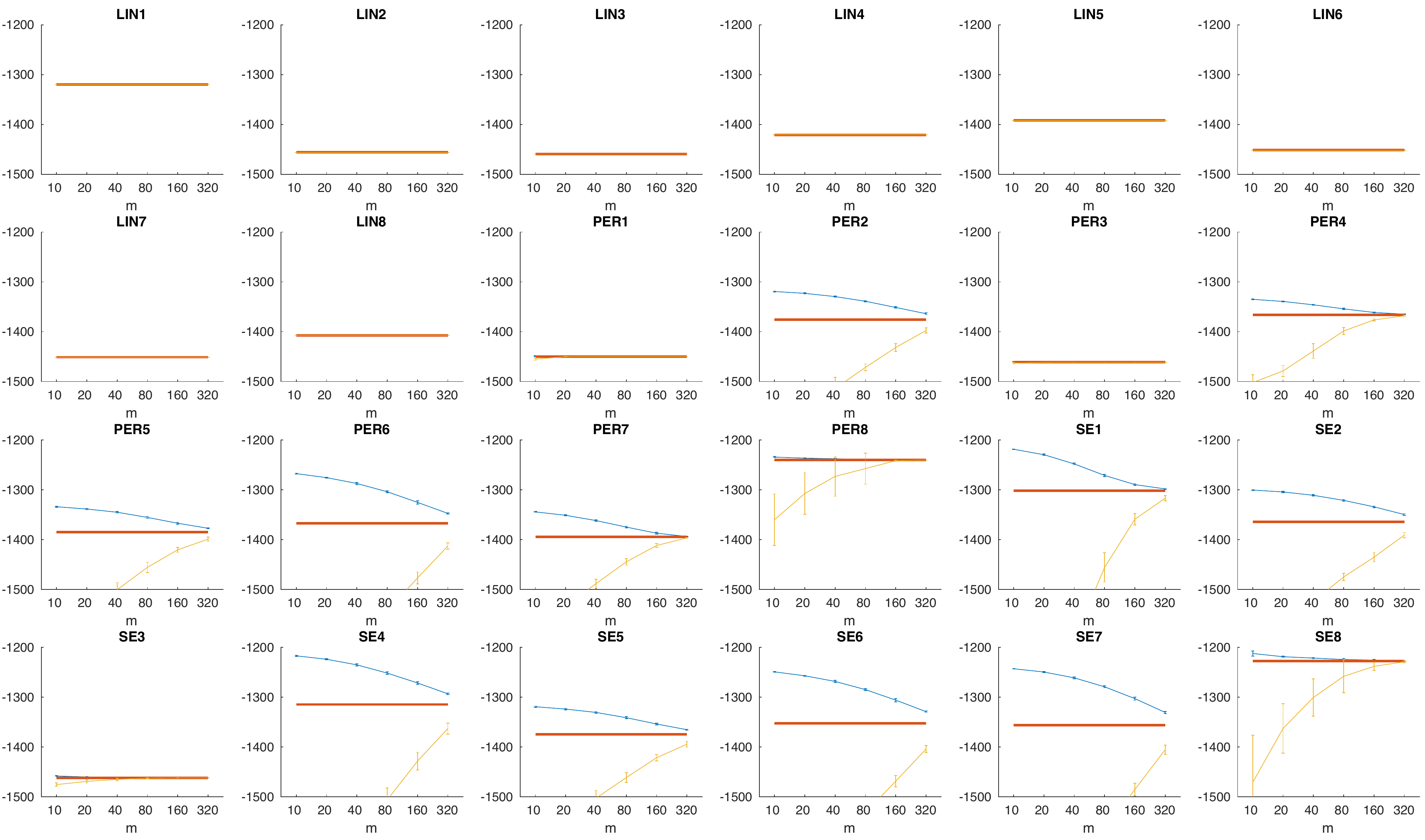}
  \caption{UB and LB for kernels at depth 1 on each dimension of concrete data, while varying the inducing points with hyperparameters fixed to the optimal values for the full GP. Error bars show mean $\pm$ 1 standard deviation over 10 random sets of inducing points.}\label{fig:vary_indpts}
\end{figure*}

Table \ref{tab:times} also compares times for the different computations after fixing the inducing points. The gains from using the variational LB instead of the full GP is clear, especially for the larger data sets, and we also confirm that it is indeed the optimisation of the LB that is the bottleneck in terms of computational cost. We also see that the NIP UB computation times are similarly fast for all $m$, thus convergence of PCG with the PIC preconditioner is happening in only a few iterations. 

\subsection{SKC on small data sets}
\label{sec:smallexp}

\begin{figure*}[h!]
  \centering
  \includegraphics[width=0.7\linewidth]{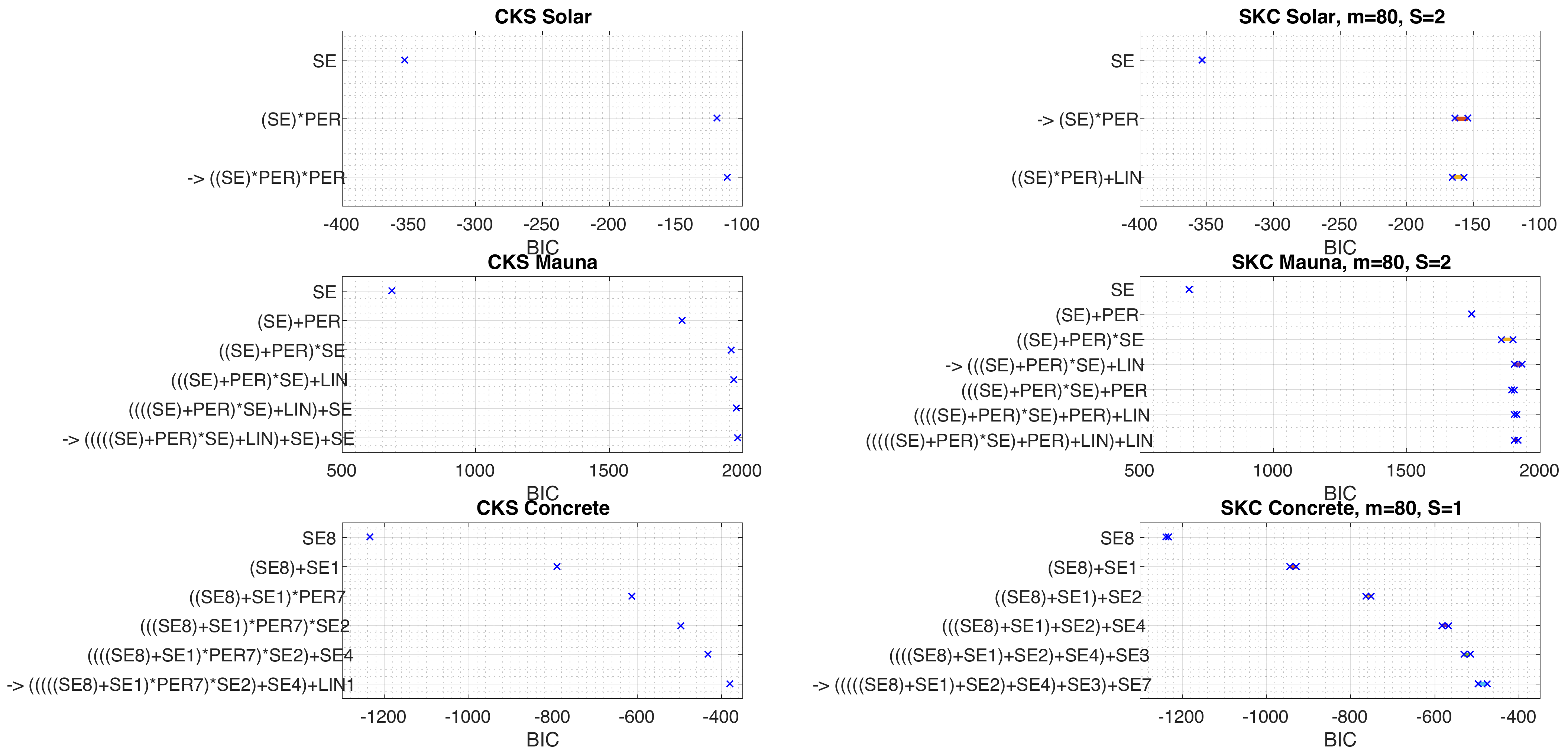}
  \caption{CKS \& SKC results for up to depth 6. Left: BIC of kernels chosen at each depth by CKS. Right: BIC intervals of kernels that have been added to the buffer by SKC with $m = 80$. The arrow indicates the kernel chosen at the end.}\label{fig:cks_skd_comparison}
  \vspace*{-4mm}
\end{figure*}

We compare the kernels chosen by CKS and by SKC for the three data sets. The results are summarised in Figure \ref{fig:cks_skd_comparison}. For solar, we see that SKC successfully finds SE $\times$ PER, which is the second highest kernel for CKS, with BIC very close to the top kernel. For mauna, SKC selects (SE + PER)$\times$ SE + LIN, which is third highest for CKS and a BIC very close to the top kernel. Looking at the values of hyperparameters in kernels PER and LIN found by SKC, 40 inducing points are sufficient for it to successfully find the correct periods and slopes in the data, reinforcing the claim that we only need a small $m$ to find good hyperparameters (see Appendix \ref{apd:mauna_solar_plots_hyps} for details). For concrete, a more challenging eight dimensional data set, we see that the kernels selected by SKC do not match those selected by CKS, but it still manages to find similar additive structure such as SE1+SE8 and SE4. Of course, the BIC intervals for kernels found by SKC are for hyperparameters found by VAR with $m = 80$, hence do not necessarily contain the optimal BIC of kernels in CKS. However the above results show that our method is still capable of selecting appropriate kernels even for low values of $m$, without having to home in to the optimal BIC using high values of $m$. The savings in computation time is significant even for these small data sets, as shown in Table \ref{tab:times}.

\subsection{SKC on medium-sized data sets \& Why the lower bound is not enough}
\label{sec:largeexp}

The UB has marginal benefits over the LB for small data sets, where the LB is already a fairly good estimate of the true BIC. However, this gap becomes significant as $N$ grows and as kernels become more complex; the UB is much tighter and more stable with respect to the choice of inducing points (as shown in Figure \ref{fig:vary_indpts}), and plays a crucial role in the model selection.

\begin{figure}[h!]
  \centering
  \subfloat{\includegraphics[width=\columnwidth]{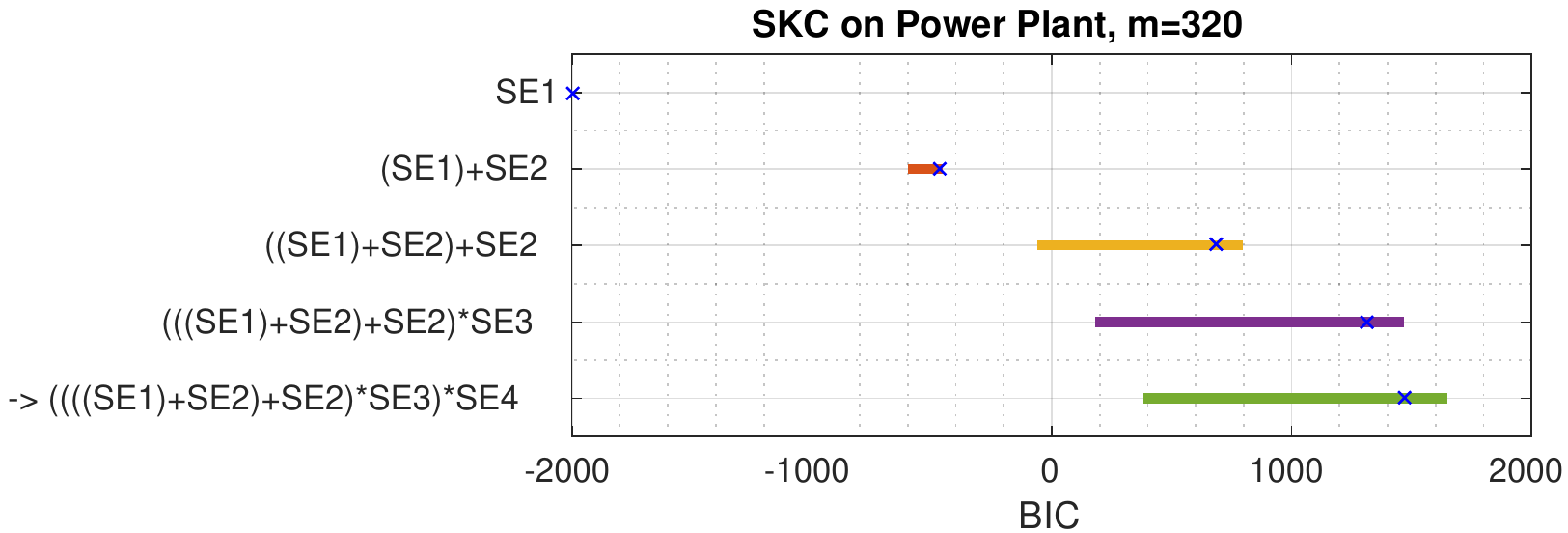}}
  \hfill
  \subfloat{\includegraphics[width=\columnwidth]{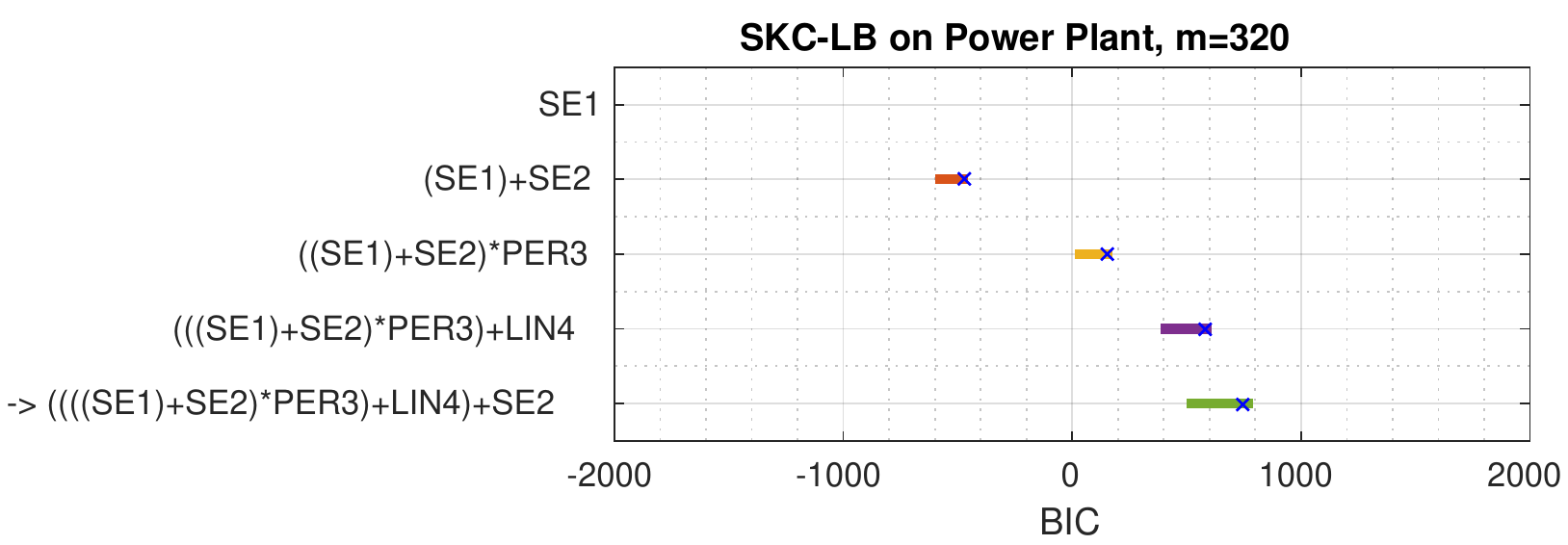}}
  \caption{Comparison of SKC (top) and SKC-LB (bottom), with $m=320, S=1$ to depth 5 on Power Plant data. The format is the same as Figure \ref{fig:cks_skd_comparison} but with the crosses at the true BIC instead of the bounds.} \label{fig:pp_skc}
  \vspace*{-3mm}
\end{figure}

\textbf{Power Plant} We first show this for SKC on the Power Plant data with 9568 observations and 4 covariates \cite{tufekci2014prediction}. We see from Figure \ref{fig:pp_skc} that again the UB is much tighter than the LB, especially in more complex kernels further down the search tree. Comparing the two plots, we also see that the use of the UB in SKC has a significant positive impact on model selection: the kernel found by SKC at depth 5 has exact BIC 1469.6, whereas the corresponding kernel found by just using the LB (SKC-LB) has exact BIC 745.5. The pathological behaviour of SKC-LB occurs at depth 3, where the (SE1+SE2)*PER3 kernel found by SKC-LB has a higher LB than the corresponding SE1+SE2+SE2 kernel found by SKC. SKC-LB chooses the former kernel since it has a higher LB, which is a failure mode since the latter kernel has a significantly higher exact BIC. Due to the tightness of the UB, we see that SKC correctly chooses the latter kernel over the former, escaping the failure mode.

\textbf{CKS vs SKC runtime} We run CKS and SKC for $m=160,320,640$ on Power Plant data on the same machine with the same hyperparameter setting. The runtimes up to depths 1/2 are: 28.7h/94.7h (CKS), 0.6h/2.6h (SKC, $m=160$), 1.1h/4.1h (SKC, $m=320$), 4.2h/15.0h (SKC, $m=640$). Again, the reduction in computation time for SKC are substantial.

\textbf{Time Series} We also implement SKC on medium-sized time series data to explore whether it can successfully find known structure at different scales. Such data sets occur frequently for hourly time series over several years or daily time series over centuries. 

\textbf{GEFCOM} First we use the energy load data set from the 2012 Global Energy Forecasting Competition \cite{hong2014global}, and hourly time series of energy load from 2004/01/01 to 2008/06/30, with 8 weeks of missing data, giving $N=38,070$. Notice that this data size is far beyond the scope of full GP optimisation in CKS.

\begin{figure}[h!]
  \vspace*{-2mm}
  \centering
  \includegraphics[width=0.8\columnwidth]{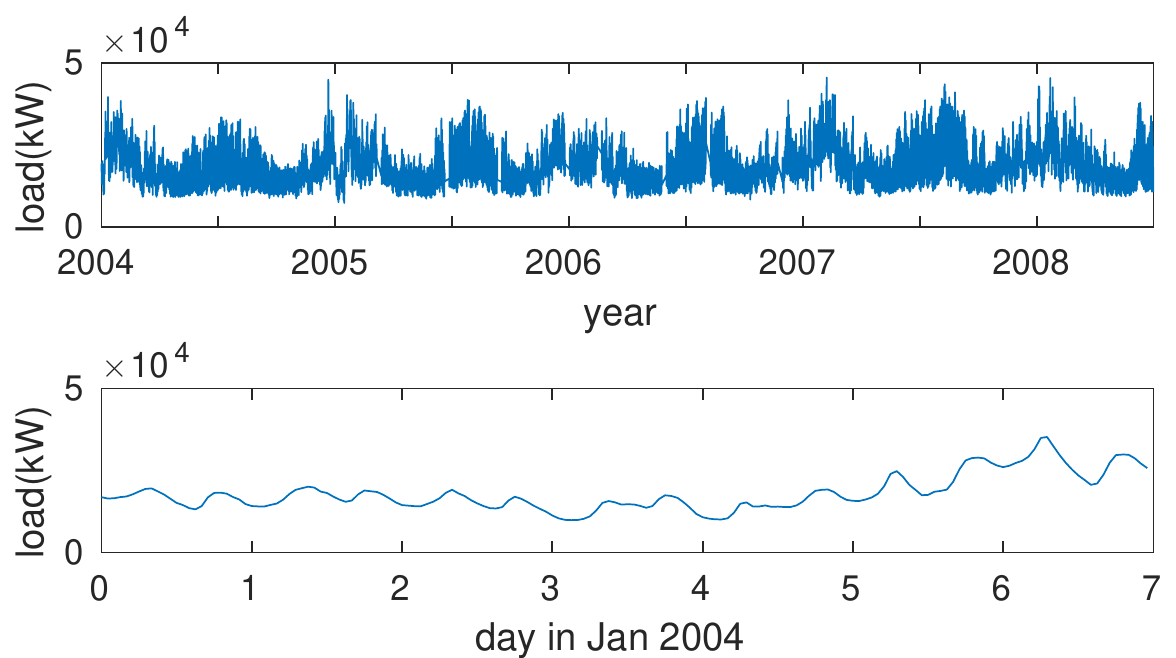}
  \caption{Top: plot of full GEFCOM data. Bottom: zoom in on the first 7 days.}\label{fig:gefcom}
  \vspace*{-2mm}
\end{figure}

% \begin{figure*}[h!]
%   \centering
%   \subfloat{\includegraphics[width=0.7\textwidth]{gefcom_zone1_data_plot}\label{fig:gefcom}}
%   \hfill
%   \subfloat{\includegraphics[width=0.7\textwidth]{tidal_plot}\label{fig:tidal}}
%   \caption{Top left: Plot of full GEFCOM data. Top right: Zoom in for the first 7 days. Bottom left: Plot of full tidal data. Bottom right: Zoom in for the first 4 weeks.}
% \end{figure*}

From the plots, we can see that there is a noisy 6-month periodicity in the time series, as well as a clear daily periodicity with peaks in the morning and evening. Despite the periodicity, there are some noticeable irregularities in the daily pattern.

\begin{table}[h!]
\vspace*{-1mm}
\caption{Hyperparameters of kernels found by SKC on GEFCOM data and on Tidal data after normalising $y$. Length scales, periods, and location converted to original scale, $\sigma^2$ left as was found with normalised $y$.}
\label{tab:skc_gefcom_tidal}
\vspace*{-3mm}
\begin{center}
\resizebox{0.8\columnwidth}{!}{
\begin{tabular}{|l|l|l|l|}
\hline
% \textbf{GEFCOM} &{} &\textbf{Tidal} &{} \\ 
\multicolumn{2}{|c|}{\textbf{GEFCOM}} & \multicolumn{2}{|c|}{\textbf{Tidal}}  \\
\hline
SE\textsubscript{1}  &$\sigma^2=0.44$ &SE                   &$\sigma^2=2.32$  \\
                     &$l=60.5$ days   &                     &$l=82.5$ days \\
\hline
PER\textsubscript{1} &$\sigma^2=1.10$ &PER\textsubscript{1} &$\sigma^2=5.17$ \\
                     &$l=1089$ days   &                     &$l=1026$ days  \\
                     &$p=1.003$ days  &                     &$p=0.538$ days \\
\hline
SE\textsubscript{2}  &$\sigma^2=0.18$ &LIN                  &$\sigma^2=0.18$ \\
                     &$l=331$ days    &                     &$loc=$ year $2015.0$  \\
\hline
PER\textsubscript{2} &$\sigma^2=0.06$ &PER\textsubscript{2} &$\sigma^2=0.08$ \\
                     &$l=170$ days    &                     &$l=5974$ days  \\
                     &$p=174$ days    &                     &$p=0.500$ days \\
\hline
LIN                  &$\sigma^2=0.17$ &PER\textsubscript{3} &$\sigma^2=0.21$ \\
                     &$loc=$ year $2006.1$  &               &$l=338$ days  \\
                     &                      &               &$p=14.6$ days \\
\hline
\end{tabular}
}
\end{center}
\vspace*{-4mm}
\end{table}

The SE\textsubscript{1} $\times$ PER\textsubscript{1}+ SE\textsubscript{2} $\times$ (PER\textsubscript{2} + LIN) kernel found by SKC with $m=160$ and its hyperparameters are summarised in the first two columns of Table \ref{tab:skc_gefcom_tidal}. Note that with only 160 inducing points, SKC has succesfully found the daily periodicity (PER\textsubscript{1}) and the 6-month periodicity (PER\textsubscript{2}). The SE\textsubscript{1} kernel in the first additive term suggests a local periodicity, exhibiting the irregular observation pattern that is repeated daily. Also note that the hyperparameter corresponding to the longer periodicity is close but not exactly half a year, owing to the noisiness of the periodicity that is apparent in the data. Moreover the magnitude of the second additive term SE\textsubscript{2} $\times$ PER\textsubscript{2} that contains the longer periodicity is 0.18 $\times$ 0.06, which is much smaller than the magnitude 0.44 $\times$ 1.10 of the first additive term SE\textsubscript{1} $\times$ PER\textsubscript{1}. This explicitly shows that the second term has less effect than the first term on the behaviour of the time series, hence the weaker long-term periodicity.

Running the kernel search just using the LB with the same hyperparameters as SKC, the algorithm is only able to detect the daily periodicity and misses the 6-month periodicity. This is consistent with the observation that the LB is loose compared to the UB, especially for bigger data sets, hence fails to evaluate the kernels correctly. We also tried selecting a random subset of the data (sizes $320,640,1280,2560$) and running CKS. For even a subset of size $2560$, the algorithm could not find the daily periodicity. None of the four subset size values were able to make it past depth 4 in the kernel search, struggling to find sophisticated structure.

\textbf{Tidal} We also run SKC on tidal data, namely the sea level measurements in Dover from 2013/01/03 to 2016/08/31 \cite{tidal_data}. This is an hourly time series, with each observation taken to be the mean of four readings taken every 15 minutes, giving $N=31,957$.

\begin{figure}[h!]
  \vspace*{-2mm}
  \centering
  \includegraphics[width=0.8\columnwidth]{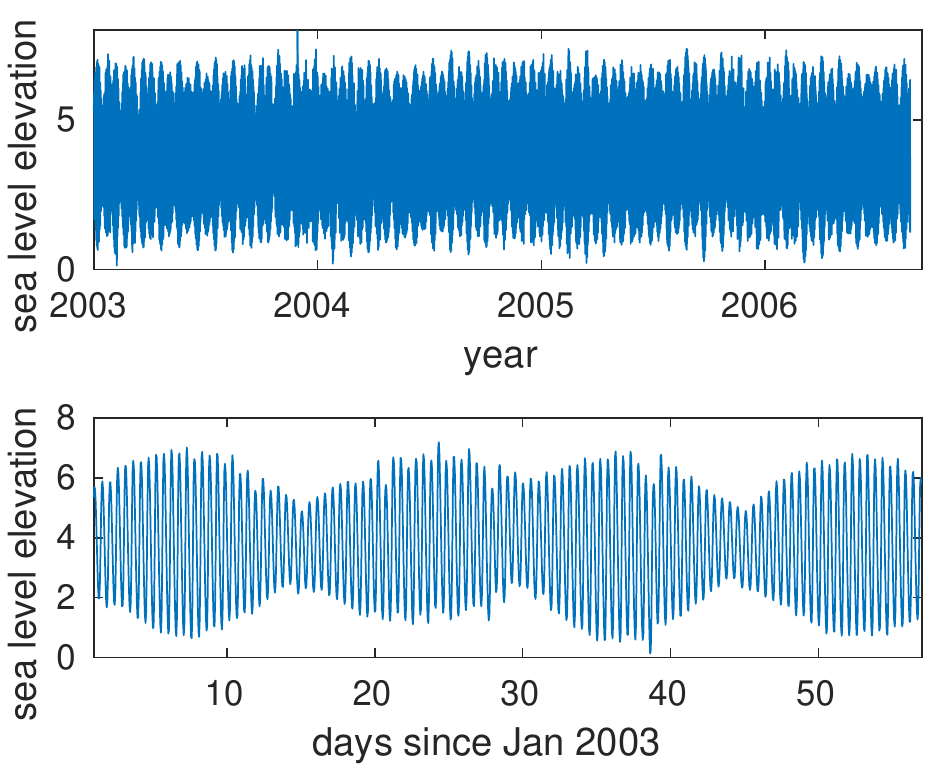}
  \caption{Top: plot of full tidal data. Right: zoom in on the first 4 weeks.}\label{fig:tidal}
  \vspace*{-3mm}
\end{figure}

Looking at the bottom of Figure \ref{fig:tidal}, we can find clear 12-hour periodicities and amplitudes that follow slightly noisy bi-weekly periodicities. The shorter periods, called semi-diurnal tides, are on average 12 hours 25 minutes $\approx$ 0.518 days long, and the bi-weekly periods arise due to gravitational effects of the moon \cite{tidal}. 

The SE $\times$ PER\textsubscript{1} $\times$ LIN $\times$ PER\textsubscript{2} $\times$ PER\textsubscript{3} kernel found by SKC with $m=640$ and its hyperparameters are summarised in the right two columns of Table \ref{tab:skc_gefcom_tidal}. The PER\textsubscript{1} $\times$ PER\textsubscript{3} kernel precisely corresponds to the doubly periodic structure in the data, whereby we have semi-daily periodicities with bi-weekly periodic amplitudes. The SE kernel has a length scale of 82.5 days, giving a local periodicity. This represents the irregularity of the amplitudes as can be seen on the bottom plot of Figure \ref{fig:tidal} between days 20 and 40. The LIN kernel has a large magnitude, but its effect is negligible when multiplied since the slope is calculated to be $-5 \times 10^{-6}$ (see Appendix \ref{apd:mauna_solar_plots_hyps} for the slope calculation formula). It essentially has the role of raising the magnitude of the resulting kernel and hence representing noise in the data. The PER\textsubscript{2} kernel is also negligible due to its high length scale and small magnitude, indicating that the amplitude of the periodicity due to this term is very small. Hence the term has minimal effect on the kernel.

The kernel search using just the LB fails to proceed past depth 3, since it is unable to find a kernel with a higher LB on the BIC than the previous depth (so the extra penalty incurred by increasing model complexity outweighs the increase in LB of log marginal likelihood). CKS on a random subset of the data similarly fails, the kernel search halting at depth 2 even for random subsets as large as size $2560$.

In both cases, SKC is able to detect the structure of data and provide accurate numerical estimates of its features, all with much fewer inducing points than $N$, whereas the kernel search using just the LB or a random subset of the data both fail.

\section{Conclusion and Discussion}
We have introduced SKC, a scalable kernel discovery algorithm that extends CKS and hence ABCD to bigger data sets. We have also derived a novel cheap upper bound to the GP marginal likelihood that sandwiches the marginal likelihood with the variational lower bound \cite{titsias2009variational}, and use this interval in SKC for selecting between different kernels. The reasons for using an upper bound instead of just the lower bound for model selection are as follows: the upper bound allows for a semi-greedy approach, allowing us to explore a wider range of kernels and compensates for the suboptimality coming from local optima in the hyperparameter optimisation. Should we wish to restrict the range of kernels explored for computational efficiency, the upper bound is tighter and more stable than the lower bound, hence we may use the upper bound as a reliable tie-breaker for kernels with overlapping intervals. Equipped with this upper bound, our method can pinpoint global/local periodicities and linear trends in time series with tens of thousands of data points, which are well beyond the reach of its predecessor CKS.

For future work we wish to make the algorithm even more scalable: for large data sets where quadratic runtime is infeasible, we can apply stochastic variational inference for GPs \cite{hensman2013gaussian} to optimise the lower bound, the bottleneck of SKC, using mini-batches of data. Finding an upper bound that is cheap and tight enough for model selection would pose a challenge. Also one drawback of the upper bound that could perhaps be resolved is that hyperparameter tuning by optimising the lower bound is difficult (see Appendix \ref{apd:analub}). One other minor scope for future work is using more accurate estimates of the model evidence than BIC. A related paper uses Laplace approximation instead of BIC for kernel evaluation in a similar kernel search context \cite{malkomes2016bayesian}. However Laplace approximation adds on an expensive Hessian term, for which it is unclear how one can obtain lower and upper bounds.

\section*{Acknowledgements}
HK and YWT’s research leading to these results has received funding from the European Research Council under the European Union’s Seventh Framework Programme (FP7/2007-2013) ERC grant agreement no. 617071. We would also like to thank Michalis Titsias for helpful discussions. 

\bibliographystyle{plain}
{\small
\bibliography{refs}}

\begin{thebibliography}{10}

\bibitem{tidal_data}
{British Oceanographic Data Centre, UK Tide Gauge Network}.
\newblock
  \url{https://www.bodc.ac.uk/data/hosted_data_systems/sea_level/uk_tide_gauge_network/processed/}.

\bibitem{bach2004multiple}
Francis~R Bach, Gert~RG Lanckriet, and Michael~I Jordan.
\newblock Multiple kernel learning, conic duality, and the smo algorithm.
\newblock In {\em ICML}, 2004.

\bibitem{bardenet2015infdpp}
R{\'e}my Bardenet and Michalis Titsias.
\newblock Inference for determinantal point processes without spectral
  knowledge.
\newblock {\em NIPS}, 2015.

\bibitem{bauer2016understanding}
Matthias~Stephan Bauer, Mark van~der Wilk, and Carl~Edward Rasmussen.
\newblock Understanding probabilistic sparse gaussian process approximations.
\newblock {\em NIPS}, 2016.

\bibitem{beal2003variational}
Matthew~James Beal.
\newblock {\em Variational algorithms for approximate Bayesian inference}.
\newblock PhD thesis, University College London, 2003.

\bibitem{boyd2004convex}
Stephen Boyd and Lieven Vandenberghe.
\newblock {\em Convex optimization}.
\newblock Cambridge University Press, 2004.

\bibitem{cortes2010impact}
Corinna Cortes, Mehryar Mohri, and Ameet Talwalkar.
\newblock On the impact of kernel approximation on learning accuracy.
\newblock In {\em AISTATS}, 2010.

\bibitem{cutajar_preconditioning_2016}
Kurt Cutajar, Michael~A. Osborne, John~P. Cunningham, and Maurizio Filippone.
\newblock Preconditioning kernel matrices.
\newblock {\em ICML}, 2016.

\bibitem{drineas2005nystrom}
Petros Drineas and Michael Mahoney.
\newblock On the nystr{\"o}m method for approximating a gram matrix for
  improved kernel-based learning.
\newblock {\em JMLR}, 6:2153--2175, 2005.

\bibitem{duvenaud2013structure}
David Duvenaud, James Lloyd, Roger Grosse, Joshua Tenenbaum, and Ghahramani
  Zoubin.
\newblock Structure discovery in nonparametric regression through compositional
  kernel search.
\newblock In {\em ICML}, 2013.

\bibitem{fraley2007bayesian}
Chris Fraley and Adrian~E Raftery.
\newblock Bayesian regularization for normal mixture estimation and model-based
  clustering.
\newblock {\em Journal of classification}, 24(2):155--181, 2007.

\bibitem{gardner17a}
Jacob Gardner, Chuan Guo, Kilian Weinberger, Roman Garnett, and Roger Grosse.
\newblock Discovering and exploiting additive structure for bayesian
  optimization.
\newblock In {\em AISTATS}, 2017.

\bibitem{grosse2012exploiting}
Roger~B. Grosse, Ruslan Salakhutdinov, William~T. Freeman, and Joshua~B.
  Tenenbaum.
\newblock Exploiting compositionality to explore a large space of model
  structures.
\newblock In {\em UAI}, 2012.

\bibitem{pmlr-v64-guyon_review_2016}
Isabelle Guyon, Imad Chaabane, Hugo~Jair Escalante, Sergio Escalera, Damir
  Jajetic, James~Robert Lloyd, Núria Macià, Bisakha Ray, Lukasz Romaszko,
  Michèle Sebag, Alexander Statnikov, Sébastien Treguer, and Evelyne Viegas.
\newblock A brief review of the chalearn automl challenge: Any-time any-dataset
  learning without human intervention.
\newblock In {\em Proceedings of the Workshop on Automatic Machine Learning},
  volume~64, pages 21--30, 2016.

\bibitem{hensman2013gaussian}
James Hensman, Nicolo Fusi, and Neil~D Lawrence.
\newblock Gaussian processes for big data.
\newblock {\em UAI}, 2013.

\bibitem{hong2014global}
Tao Hong, Pierre Pinson, and Shu Fan.
\newblock Global energy forecasting competition 2012, 2014.

\bibitem{ji2010bounded}
Chunlin Ji, Haige Shen, and Mike West.
\newblock Bounded approximations for marginal likelihoods.
\newblock 2010.

\bibitem{lean1995reconstruction}
Judith Lean, Juerg Beer, and Raymond~S Bradley.
\newblock Reconstruction of solar irradiance since 1610: Implications for
  climate cbange.
\newblock {\em Geophysical Research Letters}, 22(23), 1995.

\bibitem{lloyd2014automatic}
James~Robert Lloyd, David Duvenaud, Roger Grosse, Joshua Tenenbaum, and Zoubin
  Ghahramani.
\newblock Automatic construction and natural-language description of
  nonparametric regression models.
\newblock In {\em AAAI}, 2014.

\bibitem{malkomes2016bayesian}
Gustavo Malkomes, Charles Schaff, and Roman Garnett.
\newblock Bayesian optimization for automated model selection.
\newblock In {\em NIPS}, 2016.

\bibitem{matthews2015sparse}
Alexander G de~G Matthews, James Hensman, Richard~E Turner, and Zoubin
  Ghahramani.
\newblock On sparse variational methods and the kullback-leibler divergence
  between stochastic processes.
\newblock {\em AISTATS}, 2016.

\bibitem{tidal}
John Morrissey, James~L Sumich, and Deanna~R. Pinkard-Meier.
\newblock {\em Introduction To The Biology Of Marine Life}.
\newblock Jones \& Bartlett Learning, 1996.

\bibitem{murphy2012machine}
Kevin~P Murphy.
\newblock {\em Machine learning: a probabilistic perspective}.
\newblock MIT press, 2012.

\bibitem{oliva2016bayesian}
Junier~B Oliva, Avinava Dubey, Andrew~G Wilson, Barnab{\'a}s P{\'o}czos, Jeff
  Schneider, and Eric~P Xing.
\newblock Bayesian nonparametric kernel-learning.
\newblock In {\em AISTATS}, 2016.

\bibitem{quinonero2005unifying}
Joaquin Quinonero-Candela and Carl~Edward Rasmussen.
\newblock A unifying view of sparse approximate gaussian process regression.
\newblock {\em JMLR}, 6:1939--1959, 2005.

\bibitem{rahimi2007random}
Ali Rahimi and Benjamin Recht.
\newblock Random features for large-scale kernel machines.
\newblock In {\em NIPS}, 2007.

\bibitem{rasmussen2006gaussian}
Carl~Edward Rasmussen and Chris Williams.
\newblock {\em Gaussian processes for machine learning}.
\newblock MIT Press, 2006.

\bibitem{rudi2015less}
Alessandro Rudi, Raffaello Camoriano, and Lorenzo Rosasco.
\newblock Less is more: Nystr{\"o}m computational regularization.
\newblock In {\em NIPS}, 2015.

\bibitem{rudin1964fourier}
Walter Rudin.
\newblock Fourier analysis on groups.
\newblock {\em AMS}, 1964.

\bibitem{samo2015generalized}
Yves-Laurent~Kom Samo and Stephen Roberts.
\newblock Generalized spectral kernels.
\newblock {\em arXiv preprint arXiv:1506.02236}, 2015.

\bibitem{schwarz1978estimating}
Gideon Schwarz et~al.
\newblock Estimating the dimension of a model.
\newblock {\em The Annals of Statistics}, 6(2):461--464, 1978.

\bibitem{seeger2003fast}
Matthias Seeger, Christopher Williams, and Neil Lawrence.
\newblock Fast forward selection to speed up sparse gaussian process
  regression.
\newblock In {\em AISTATS}, 2003.

\bibitem{shewchuk1994introduction}
Jonathan~Richard Shewchuk.
\newblock An introduction to the conjugate gradient method without the
  agonizing pain, 1994.

\bibitem{snelson2005sparse}
Edward Snelson and Zoubin Ghahramani.
\newblock Sparse gaussian processes using pseudo-inputs.
\newblock In {\em NIPS}, 2005.

\bibitem{solin2014explicit}
Arno Solin and Simo S{\"a}rkk{\"a}.
\newblock Explicit link between periodic covariance functions and state space
  models.
\newblock {\em JMLR}, 2014.

\bibitem{mauna}
Pieter Tans and Ralph Keeling.
\newblock {NOAA/ESRL, Scripps Institution of Oceanography}.
\newblock \url{ftp://ftp.cmdl.noaa.gov/ccg/co2/trends/co2_mm_mlo.txt}.

\bibitem{titsias2009variational}
Michalis~K Titsias.
\newblock Variational learning of inducing variables in sparse gaussian
  processes.
\newblock In {\em AISTATS}, 2009.

\bibitem{tufekci2014prediction}
P{\i}nar T{\"u}fekci.
\newblock Prediction of full load electrical power output of a base load
  operated combined cycle power plant using machine learning methods.
\newblock {\em International Journal of Electrical Power \& Energy Systems},
  60:126--140, 2014.
\newblock
  \url{http://archive.ics.uci.edu/ml/datasets/combined+cycle+power+plant}.

\bibitem{vanhatalo2013gpstuff}
Jarno Vanhatalo, Jaakko Riihim{\"a}ki, Jouni Hartikainen, Pasi Jyl{\"a}nki,
  Ville Tolvanen, and Aki Vehtari.
\newblock Gpstuff: Bayesian modeling with gaussian processes.
\newblock {\em JMLR}, 14(Apr):1175--1179, 2013.

\bibitem{williams2001using}
Christopher Williams and Matthias Seeger.
\newblock Using the nystr{\"o}m method to speed up kernel machines.
\newblock In {\em NIPS}, 2001.

\bibitem{wilson2013gaussian}
Andrew~Gordon Wilson and Ryan~Prescott Adams.
\newblock Gaussian process kernels for pattern discovery and extrapolation.
\newblock {\em arXiv preprint arXiv:1302.4245}, 2013.

\bibitem{wilson2015msgp}
Andrew~Gordon Wilson, Christoph Dann, and Hannes Nickisch.
\newblock Thoughts on massively scalable {G}aussian processes.
\newblock {\em arXiv preprint arXiv:1511.01870}, 2015.
\newblock \url{http://arxiv.org/abs/1511.01870}.

\bibitem{wilson2016deep}
Andrew~Gordon Wilson, Zhiting Hu, Ruslan Salakhutdinov, and Eric~P Xing.
\newblock Deep kernel learning.
\newblock In {\em AISTATS}, 2016.

\bibitem{yeh1998modeling}
I-C Yeh.
\newblock Modeling of strength of high-performance concrete using artificial
  neural networks.
\newblock {\em Cement and Concrete research}, 28(12):1797--1808, 1998.
\newblock
  \url{https://archive.ics.uci.edu/ml/datasets/Concrete+Compressive+Strength}.

\end{thebibliography}

\clearpage

\appendix
\section*{Appendix}

\section{Bayesian Information Criterion (BIC)} \label{apd:bic}
The BIC is a model selection criterion that is the marginal likelihood with a model complexity penalty:
$$
BIC = \log p(y|\hat{\theta}) - \frac{1}{2} p \log(N)
$$
for observations $y$, number of observations $N$, maximum likelihood estimate (MLE) of model hyperparameters $\hat{\theta}$, number of hyperparameters p. It is derived as an approximation to the log model evidence $\log p(y)$.

\section{Compositional Kernel Search Algorithm}\label{apd:kernelsearch}
\begin{algorithm2e}
\caption{Compositional Kernel Search Algorithm}
\label{alg:kernelsearch}
\KwIn{data $x_1, \ldots, x_n \in \mathbb{R}^D, y_1, \ldots, y_n \in \mathbb{R}$,base kernel set $\mathcal{B}$}
\KwOut{$k$, the resulting kernel}
For each base kernel on each dimension, fit GP to data (i.e. optimise hyperparams by ML-II) and set $k$ to be kernel with highest BIC. \\
\For{depth=1:T (either fix T or repeat until BIC no longer increases)}{
  Fit GP to following kernels and set $k$ to be the one with highest BIC: \\
  (1) All kernels of form $k+B$ where $B$ is any base kernel on any dimension \\
  (2) All kernels of form $k\times B$ where $B$ is any base kernel on any dimension \\
  (3) All kernels where a base kernel in $k$ is replaced by another base kernel
}
\end{algorithm2e}

\section{Base Kernels}\label{apd:base}
\begin{align*}
\text{LIN}(x,x') &=\sigma^2 (x-l)(x'-l)\\
\text{SE}(x,x') &=\sigma^2\exp\Big(-\frac{(x-x')^2}{2l^2}\Big) \\
\text{PER}(x,x') &=\sigma^2\exp\Big(-\frac{2\sin^2(\pi(x-x')/p)}{l^2}\Big) 
\end{align*}

\section{Matrix Identities}\label{apd:matrixid}
\begin{lemma}[Woodbury's Matrix Inversion Lemma]\label{lem:woodbury}
$(A+UBV)^{-1}=A^{-1}-A^{-1}U(B^{-1}+VA^{-1}U)^{-1}VA^{-1}$
\end{lemma}

So setting $A=\sigma^2 I$ (Nystr{\"o}m) or $\sigma^2 I+ diag(K-\hat{K})$ (FIC) or $\sigma^2 I+ blockdiag(K-\hat{K})$ (PIC), $U=\Phi^T=V$, $B=I$, we get:
\begin{align*}
    (A+\Phi^\top\Phi)^{-1} = A^{-1} - A^{-1}\Phi^\top (I+\Phi A^{-1}\Phi^\top)^{-1}\Phi A^{-1}
\end{align*}

\begin{lemma}[Sylvester's Determinant Theorem]\label{lem:sylvester}
 $\det(I+AB)=\det(I+BA) \hspace{2mm} \forall A \in \mathbb{R}^{m\times n} \hspace{2mm}  \forall B \in \mathbb{R}^{n\times m}$
\end{lemma}

Hence: 
\begin{align*}
\det(\sigma^2 I +\Phi^\top\Phi) &=(\sigma^2)^n \det(I+\sigma^{-2}\Phi^\top\Phi) \\
&= (\sigma^2)^n \det(I+\sigma^{-2}\Phi\Phi^\top) \\
&= (\sigma^2)^{n-m} \det(\sigma^2 I+\Phi\Phi^\top) \\
\end{align*}

\section{Proof of Proposition 1} \label{apd:proof}
\begin{proof}
If PCG converges, the upper bound for NIP is exact. We showed in Section \ref{sec:lbubexp} that the convergence happened in only a few iterations. Moreover Cortes et al \cite{cortes2010impact} shows that the lower bound for NIP can be rather loose in general.

So it suffices to prove that the upper bound for NLD is tighter than the lower bound for NLD. Let $(\lambda_i)_{i=1}^N, (\hat{\lambda}_i)_{i=1}^N$ be the ordered eigenvalues of $K+\sigma^2 I, \hat{K}+\sigma^2 I$ respectively. Since $K-\hat{K}$ is positive semi-definite (e.g. \cite{bardenet2015infdpp}), we have $\lambda_i \geq \hat{\lambda_i} \geq 2\sigma^2$ $\forall i$ (using the assumption in the proposition). Now the slack in the upper bound is:
\begin{align*}
-\frac{1}{2}&\log \det (\hat{K}+\sigma^2 I) - ( - \frac{1}{2}\log \det (K+\sigma^2 I)) \\
&= \frac{1}{2}\sum_{i=1}^N (\log \lambda_i - \log \hat{\lambda}_i )
\end{align*}
Hence the slack in the lower bound is:
\begin{align*}
-\frac{1}{2}&\log \det (K+\sigma^2 I) \\
& - \Big[-\frac{1}{2}\log \det (\hat{K}+\sigma^2 I) - \frac{1}{2\sigma^2}\Tr(K-\hat{K})\Big] \\
& = - \frac{1}{2}\sum_{i=1}^N (\log \lambda_i - \log \hat{\lambda}_i) +\frac{1}{2\sigma^2} \sum_{i=1}^N (\lambda_i - \hat{\lambda}_i)
\end{align*}
Now by concavity and monotonicity of $\log$, and since $\hat{\lambda} \geq 2\sigma^2$, we have:
\begin{align*}
    \frac{\log \lambda_i - \log \hat{\lambda}_i}{\lambda_i - \hat{\lambda}_i} & \leq \frac{1}{2\sigma^2} \\
    \Rightarrow \sum_{i=1}^N (\log \lambda_i - \log \hat{\lambda}_i) &\leq \frac{1}{2\sigma^2} \sum_{i=1}^N (\lambda_i - \hat{\lambda}_i) \\
    \Rightarrow \frac{1}{2} \sum_{i=1}^N (\log \lambda_i - \log \hat{\lambda}_i) & \\
    \leq \frac{1}{2\sigma^2} \sum_{i=1}^N (\lambda_i - \hat{\lambda}_i) & - \frac{1}{2} \sum_{i=1}^N (\log \lambda_i - \log \hat{\lambda}_i) \hspace{4mm} 
\end{align*}
\end{proof}

\begin{figure*}
  \centering
  \includegraphics[width=0.9\linewidth]{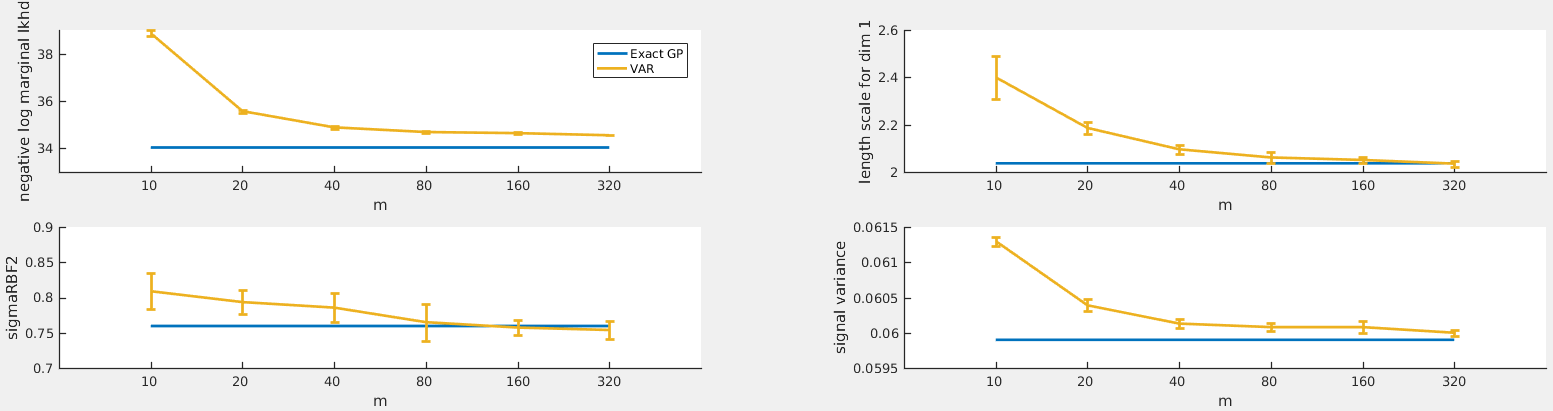}
  \caption{Log marginal likelihood and hyperparameter values after optimising the lower bound with ARD kernel on a subset of the Power plant data for different values of $m$. This is compared against the exact GP values when optimising the true log marginal likelihood. Error bars show mean $\pm$ 1 standard deviation over 10 random iterations.}\label{fig:pp500}
\end{figure*}

\begin{figure*}
  \centering
  \includegraphics[width=0.7\linewidth]{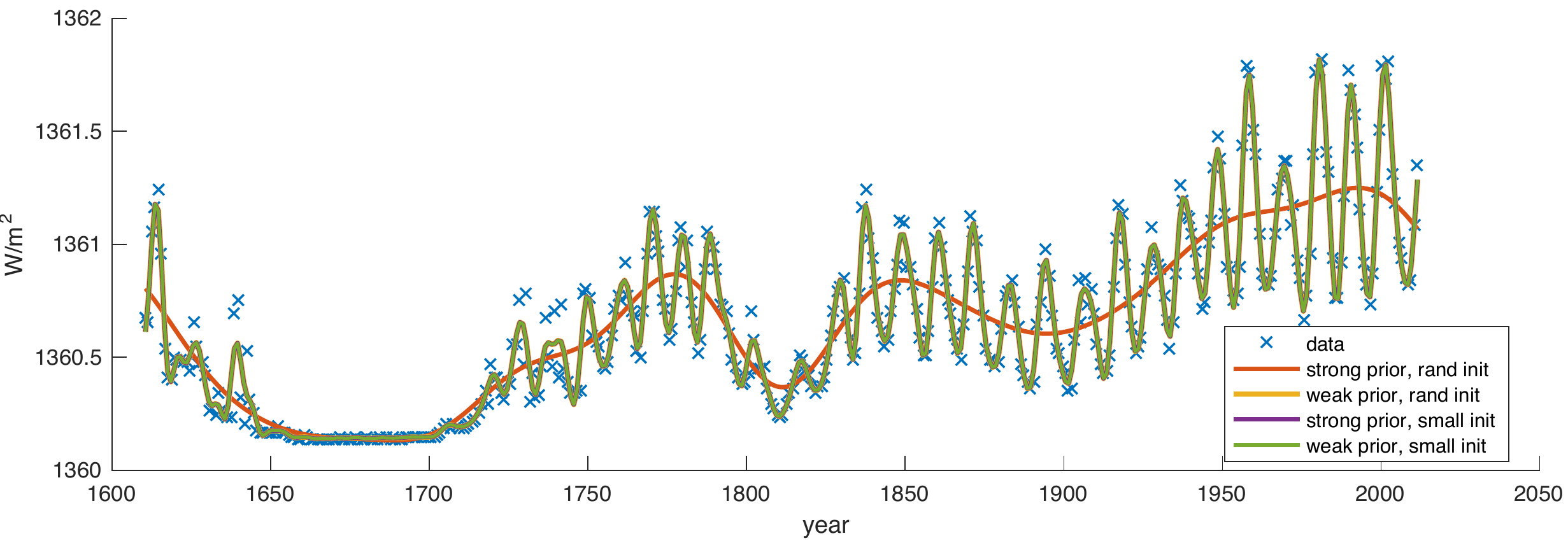}
  \caption{GP predictions on solar data set with SE kernel for different priors and initialisations.}\label{fig:prior_and_init}
\end{figure*}

\section{Convergence of hyperparameters from optimising lower bound to optimal hyperparameters} \label{apd:hyp_conv}

Note from Figure \ref{fig:pp500} that the hyperparameters found by optimising the lower bound converges to the hyperparameters found by the exact GP when optimising the exact marginal likelihood, giving empirical evidence for the second claim in Section \ref{subsec:skc}.

\section{Parallelising SKC} \label{apd:parallel}
Note that SKC can be parallelised across the random hyperparameter initialisations, and also across the kernels at each depth for computing the BIC intervals. In fact, SKC is even more parallelisable with the kernel buffer: say at a certain depth, we have two kernels remaining to be optimised and evaluated before we can move onto the next depth. If the buffer size is 5, say, then we can in fact move on to the next depth and grow the kernel search tree on the top 3 kernels of the buffer, without having to wait for the 2 kernel evaluations to be complete. This saves a lot of computation time wasted by idle cores waiting for all kernel evaluations to finish before moving on to the next depth of the kernel search tree.

\section{Optimisation}
\label{apd:opt}
Since we wish to use the learned kernels for interpretation, it is important to have the hyperparameters lie in a sensible region after the optimisation. In other words, we wish to regularise the hyperparameters during optimisation. For example, we want the SE kernel to learn a globally smooth function with local variation. When na{\"i}vely optimising the lower bound, sometimes the length scale and the signal variance becomes very small, so the SE kernel explains all the variation in the signal and ends up connecting the dots. We wish to avoid this type of behaviour. This can be achieved by giving priors to hyperparameters and optimising the energy (log prior added to the log marginal likelihood) instead, as well as using sensible initialisations. Looking at Figure \ref{fig:prior_and_init}, we see that using a strong prior with a sensible random initialisation (see Appendix \ref{apd:hyp_init_priors} for details) gives a sensible smoothly varying function, whereas for all the three other cases, we have the length scale and signal variance shrinking to small values, causing the GP to overfit to the data. Note that the weak prior is the default prior used in the GPstuff software \cite{vanhatalo2013gpstuff}.

Careful initialisation of hyperparameters and inducing points is also very important, and can have strong influence the resulting optima. It is sensible to have the optimised hyperparameters of the parent kernel in the search tree be inherited and used to initialise the hyperparameters of the child. The new hyperparameters of the child must be initialised with random restarts, where the variance is small enough to ensure that they lie in a sensible region, but large enough to explore a good portion of this region. As for the inducing points, we want to spread them out to capture both local and global structure. Trying both K-means and a random subset of training data, we conclude that they give similar results and resort to a random subset. Moreover we also have the option of learning the inducing points. However, this will be considerably more costly and show little improvement over fixing them, as we show in Section \ref{sec:exp}. Hence we do not learn the inducing points, but fix them to a given randomly chosen set. 

Hence for SKC, we use \textit{maximum a posteriori} (MAP) estimates instead of MLE for the hyperparameters to calculate the BIC, since the priors have a noticeable effect on the optimisation. This is justified \cite{murphy2012machine} and has been used for example in \cite{fraley2007bayesian}, where they argue that using the MLE to estimate the BIC for Gaussian mixture models can fail due to singularities and degeneracies.

\section{Hyperparameter initialisation and priors}
\label{apd:hyp_init_priors}
$Z \sim \mathcal{N}(0,1), TN(\sigma^2,I)$ is a Gaussian with mean 0 and variance $\sigma^2$ truncated at the interval $I$ then renormalised.\\
\textbf{Signal noise} \\
$\sigma^2 = 0.1 \times \exp (Z/2)$ \\
$p(\log \sigma^2) = \mathcal{N}(0,0.2)$ \\
\\
\textbf{LIN} \\
$\sigma^2 = \exp (V)$ where $V \sim TN(1,[-\infty,0]), l = \exp (\frac{Z}{2})$ \\
$p(\log \sigma^2) = logunif$ \\
$p(\log l) = logunif$ \\
\\
\textbf{SE} \\
$l = \exp (Z/2), \sigma^2 = 0.1 \times \exp (Z/2)$ \\
$p(\log l) = \mathcal{N}(0,0.01), p(\log \sigma^2) = logunif$ \\
\\
\textbf{PER} \\
$p_{min} = \log (10 \times \frac{\max(x)-\min(x)}{N})$ (shortest possible period is 10 time steps) \\
$p_{max} = \log (\frac{\max(x)-\min(x)}{5})$ (longest possible period is a fifth of the range of data set) \\
$l = \exp(Z/2), p = \exp(p_{min} + W)$ or $\exp(p_{max} + U)$, $\sigma^2 = 0.1 \times \exp(Z/2)$ w.p. $\frac{1}{2}$\\
where $W \sim \mathcal{TN}(-0.5,[0,\infty])$, \\
$U \sim \mathcal{TN}(-0.5,[-\infty,0]))$\\
$p(\log l) = t(\mu = 0, \sigma^2 = 1, \nu = 4)$, \\
$p(\log p) = \mathcal{LN}(p_{min} - 0.5,0.25)$ or $\mathcal{LN}(p_{max} - 2,0.5)$ w.p. $\frac{1}{2}$ \\
$p(\log \sigma^2) = logunif$ where $\mathcal{LN}(\mu,\sigma^2)$ is log Gaussian, $t(\mu, \sigma^2, \nu)$ is the student's t-distribution.

\section{Computation times} \label{apd:times}
\begin{table*}
\caption{Mean and standard deviation of computation times (in seconds) for full GP optimisation, Var GP optimisation(with and without learning inducing points), NLD and NIP (PCG using PIC preconditioner) upper bounds over 10 random iterations.}
\label{tab:times}
\begin{center}
\resizebox{0.7\linewidth}{!}{
\begin{tabular}{|l|l|l|l|}
\hline
{}                  &{\textbf{Solar}}   &{\textbf{Mauna}}   &{\textbf{Concrete}}\\ 
\hline
\textbf{GP}                  &$29.1950 \pm 5.1430$    &$164.8828 \pm 58.7865$    &$403.8233
 \pm 127.0364$\\ 
\hline
\textbf{Var GP}\hspace{4.5mm} m=10        &$7.0259 \pm 4.3928 $    &$6.0117 \pm  3.8267$    &$5.4358 \pm 0.7298 $\\
\hspace{16mm} m=20           &$8.3121 \pm 5.4763$    &$11.9245 \pm  6.8790$    &$10.2410 \pm 2.5109$\\
\hspace{16mm} m=40           &$10.2263\pm 4.1025$    &$17.1479 \pm  10.7898$    &$19.6678 \pm 4.3924$\\
\hspace{16mm} m=80           &$9.6752 \pm 6.5343$    &$28.9876  \pm 13.0031$    &$47.2225 \pm 13.1955$\\
\hspace{16mm} m=160          &$25.6330 \pm 8.7934$    &$91.0406 \pm 39.8409 $    &$158.9199 \pm 18.1276$\\
\hspace{16mm} m=320          &$76.3447 \pm 20.3337$    &$202.2369 \pm 96.0749$    &$541.4835 \pm 99.6145$\\
\hline
\textbf{NLD}\hspace{8.5mm} m=10           &$0.0019 \pm 0.0001$    &$0.0033 \pm 0.0002$    &$0.0113\pm 0.0004$\\
\hspace{16mm} m=20           &$0.0026 \pm 0.0001$    &$0.0046 \pm 0.0002$    &$0.0166\pm  0.0007 $\\
\hspace{16mm} m=40           &$0.0043 \pm 0.0001$    &$0.0079 \pm 0.0003$    &$0.0286 \pm 0.0005$\\
\hspace{16mm} m=80           &$0.0084 \pm 0.0002$    &$0.0154 \pm 0.0004$    &$0.0554 \pm  0.0012$\\
\hspace{16mm} m=160          &$0.0188 \pm 0.0006$    &$0.0338 \pm 0.0007$    &$0.1188  \pm 0.0030$\\
\hspace{16mm} m=320          &$0.0464 \pm 0.0032$    &$0.0789 \pm 0.0036$    &$0.2550 \pm 0.0074$\\
\hline
\textbf{NIP} \hspace{9mm} m=10           &$0.0474 \pm 0.0092$    &$0.1020  \pm 0.0296$    &$0.2342  \pm 0.0206$\\
\hspace{16mm} m=20           &$0.0422 \pm 0.0130$    &$0.1274 \pm  0.0674$    &$0.1746  \pm 0.0450$\\
\hspace{16mm} m=40           &$0.0284 \pm 0.0075$    &$ 0.0846 \pm 0.0430$    &$0.2345 \pm 0.0483$\\
\hspace{16mm} m=80           &$0.0199 \pm 0.0081$    &$0.0553 \pm 0.0250$    &$0.2176  \pm 0.0376$\\
\hspace{16mm} m=160          &$0.0206 \pm 0.0053$    &$0.0432  \pm 0.0109$    &$0.2136  \pm 0.0422$\\
\hspace{16mm} m=320          &$0.0250 \pm 0.0019$    &$0.0676 \pm 0.0668$    &$0.2295 \pm 0.0433$\\   
\hline
\textbf{Var GP},\hspace{4mm} m=10           &$23.4 \pm 14.6$    &$42.0  \pm 33.0 $   &$110.0  \pm 302.5$\\
\textbf{learn IP} \hspace{3mm} m=20           &$38.5 \pm 17.5$    &$62.0 \pm  66.0$    &$70.0  \pm 97.0$\\
\hspace{16mm} m=40           &$124.7 \pm 99.0$    &$ 320.0 \pm 236.0   $ &$307.0 \pm 341.4$\\
\hspace{16mm} m=80           &$268.6 \pm 196.6$    &$1935.0 \pm 1103.0   $ &$666.0  \pm 41.0$\\
\hspace{16mm} m=160          &$1483.6 \pm 773.8$    &$10480.0  \pm 5991.0$    &$4786.0  \pm 406.9$\\
\hspace{16mm} m=320          &$2923.8 \pm 1573.5$    &$39789.0 \pm 23870.0$    &$25906.0 \pm 820.9$\\
\hline
\end{tabular}
}
\end{center}
\end{table*}

Look at Table \ref{tab:times}. For all three data sets, the GP optimisation time is much greater than the sum of the Var GP optimisation time and the upper bound (NLD + NIP) evaluation time for $m \leq 80$. Hence the savings in computation time for SKC is significant even for these small data sets.

Note that we show the lower bound optimisation time against the upper bound evaluation time instead of the evaluation times for both, since this is what happens in SKC - the lower bound has to be optimised for each kernel, whereas the upper bound only has to be evaluated once.

\section{Mauna and Solar plots and hyperparameter values found by SKC}
\label{apd:mauna_solar_plots_hyps}

\begin{figure}[h!]
  \centering
  \subfloat[Solar]{\includegraphics[width=0.8\columnwidth]{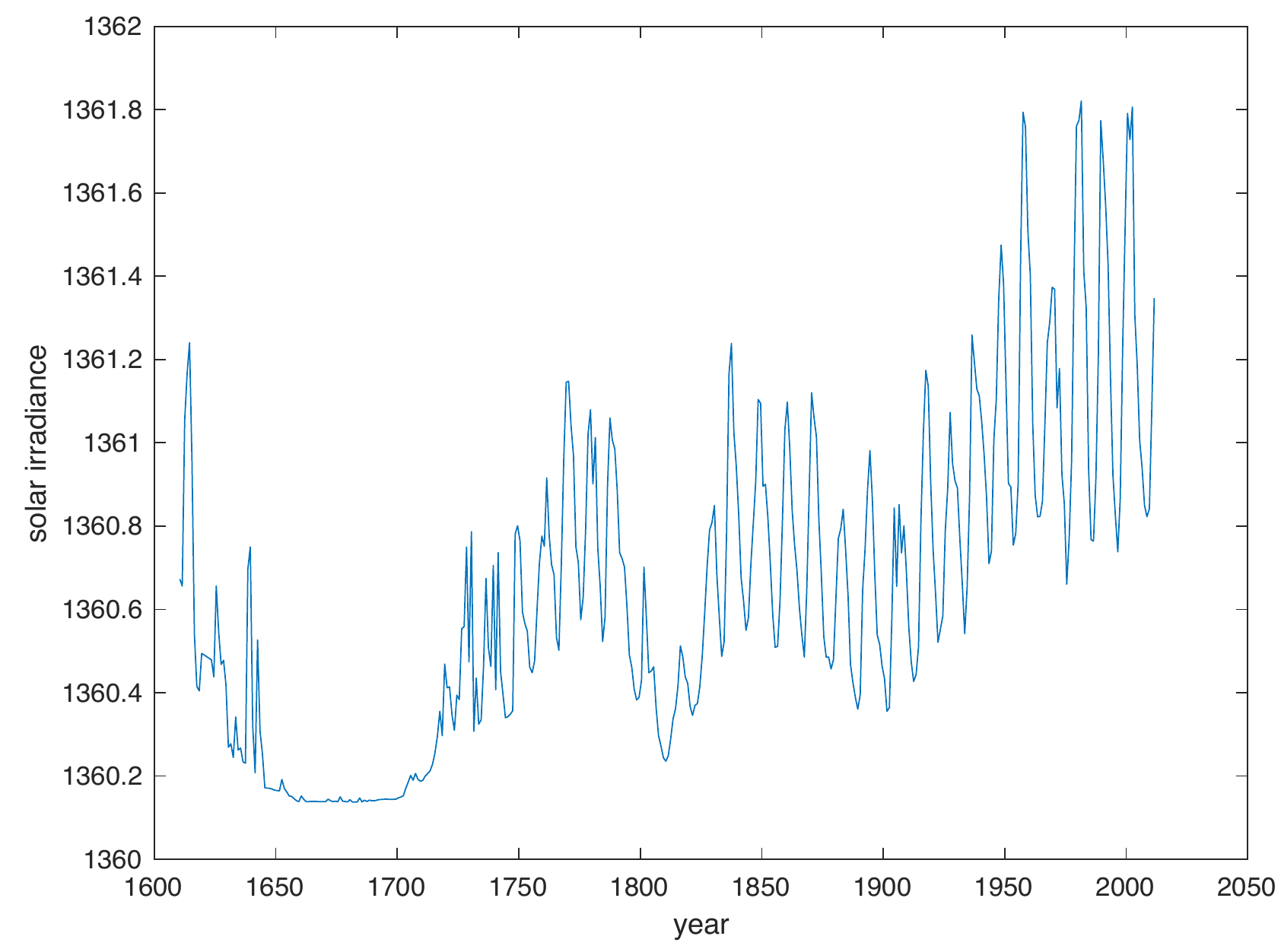}\label{fig:solar}}
  \hfill
  \subfloat[Mauna]{\includegraphics[width=0.8\columnwidth]{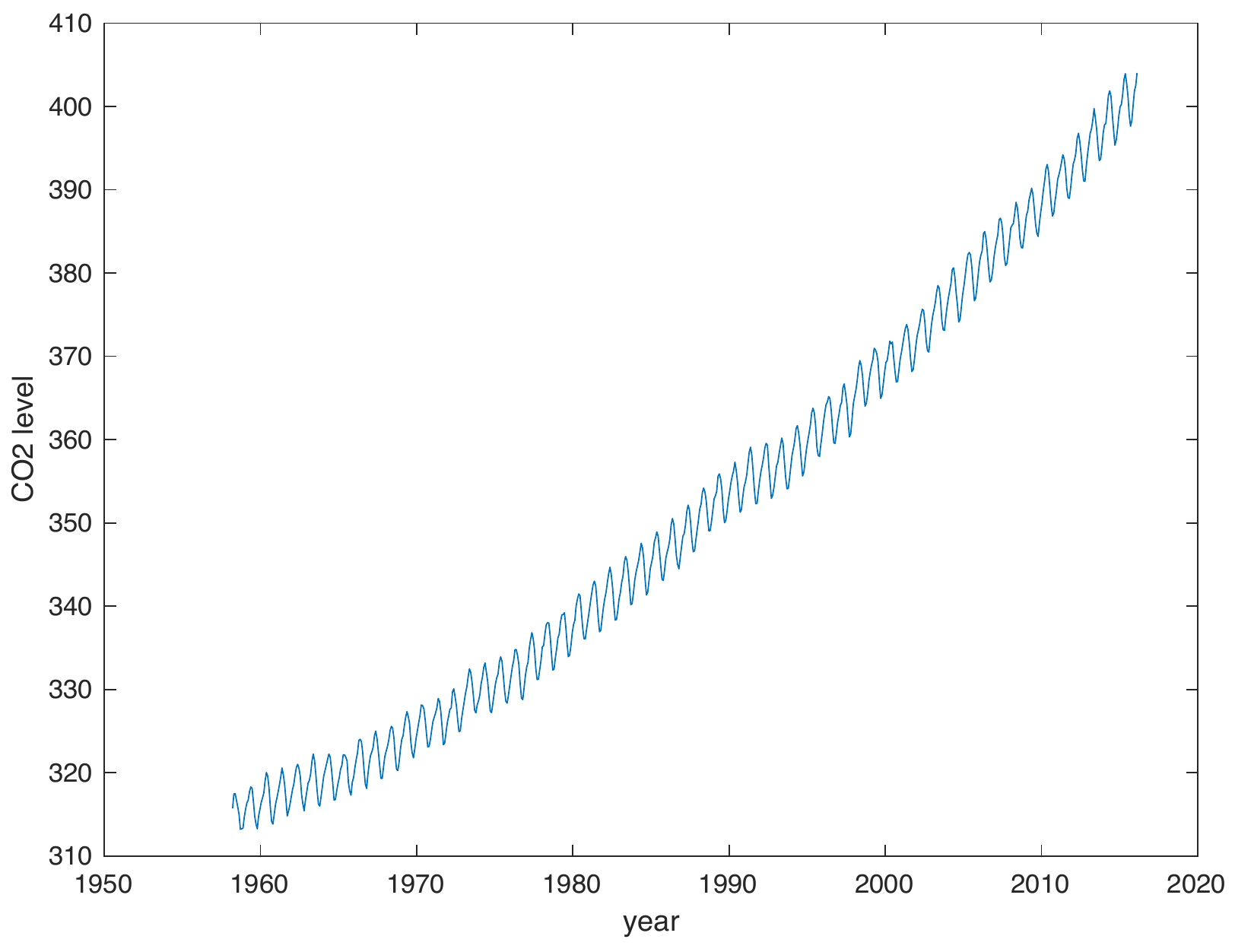}\label{fig:mauna}}
  \caption{Plots of small time series data: Solar and Mauna}
\end{figure}

\paragraph{Solar} The solar data has 26 cycles over 285 years, which gives a periodicity of around 10.9615 years. Using SKC with $m=40$, we find the kernel: SE $\times$ PER $\times$ SE $\equiv$ SE $\times$ PER. The value of the period hyperparameter in PER is 10.9569 years, hence SKC finds the periodicity to 3 s.f. with only 40 inducing points. The SE term converts the global periodicity to local periodicity, with the extent of the locality governed by the length scale parameter in SE, equal to 45. This is fairly large, but smaller than the range of the domain (1610-2011), indicating that the periodicity spans over a long time but isn't quite global. This is most likely due to the static solar irradiance between the years 1650-1700, adding a bit of noise to the periodicities.

\paragraph{Mauna} 
The annual periodicity in the data and the linear trend with positive slope is clear. Linear regression gives us a slope of 1.5149. SKC with $m=40$ gives the kernel: SE + PER + LIN. The period hyperparameter in PER takes value 1, hence SKC successfully finds the right periodicity. The offset $l$ and magnitude $\sigma^2$ parameters of LIN allow us to calculate the slope by the formula $\sigma^2 (x-l)^\top [\sigma^2(x-l)(x-l)^\top + \sigma_n^2 I]^{-1} y$ where $\sigma_n^2$ is the noise variance in the learned likelihood. This formula is obtained from the posterior mean of the GP, which is linear in the inputs for the linear kernel. This value amounts to 1.5150, hence the slope found by SKC is accurate to 3 s.f.

\section{Optimising the upper bound}
\label{apd:analub}
If the upper bound is tighter and more robust with respect to choice of inducing points, why don't we optimise the upper bound to find hyperparameters? If this were to be possible then we can maximise this to get an upper bound of the exact marginal likelihood with optimal hyperparameters. In fact this holds for any analytic upper bound whose value and gradients can be evaluated cheaply. Hence for any $m$, we can find an interval that contains the true optimised marginal likelihood. So if this interval is dominated by an interval of another kernel, we can discard the kernel and there is no need to evaluate the bounds for bigger values of $m$. Now we wish to use values of $m$ such that we can choose the right kernel (or kernels) at each depth of the search tree with minimal computation. This gives rise to an exploitation-exploration trade-off, whereby we want to balance between raising $m$ for tight intervals that allow us to discard unsuitable kernels whose intervals fall strictly below that of other kernels, and quickly moving on to the next depth in the search tree to search for finer structure in the data. The search algorithm is highly parallelisable, and thus we may raise $m$ simultaneously for all candidate kernels. At deeper levels of the search tree, there may be too many candidates for simultaneous computation, in which case we may select the ones with the highest upper bound to get tighter intervals. Such attempts are listed below.

Note the two inequalities for the NLD and NIP terms:
\begin{align}
 -\frac{1}{2}\log \det (\hat{K}+\sigma^2 I) - &\frac{1}{2\sigma^2}\Tr(K-\hat{K}) \nonumber\\
 \leq -\frac{1}{2}\log \det& (K+\sigma^2 I)  \nonumber\\
 \leq &-\frac{1}{2}\log \det (\hat{K}+\sigma^2 I) \\
 -\frac{1}{2}y^\top(\hat{K}+\sigma^2 I)^{-1}y& \nonumber\\
 \leq -\frac{1}{2}y^\top  (K+&\sigma^2 I)^{-1} y \nonumber\\
 \leq & -\frac{1}{2}y^\top(K+(\sigma^2 +\Tr(K-\hat{K}))I)^{-1}y
\end{align}
Where the first two inequalities come from \cite{bardenet2015infdpp}, the third inequality is a direct consequence of $K-\hat{K}$ being postive semi-definite, and the last inequality is from Michalis Titsias' lecture slides \footnote{http://www.aueb.gr/users/mtitsias/papers/titsiasNipsVar14.pdf}.

Also from \ref{eq:cg}, we have that 
\begin{equation*}
-\frac{1}{2}\log \det (\hat{K}+\sigma^2 I) + \frac{1}{2}\alpha^\top(K+\sigma^2 I)\alpha-\alpha^\top y
\end{equation*}
is an upper bound $\forall \alpha \in \mathbb{R}^N$. Thus one idea of obtaining a cheap upper bound to the optimised marginal likelihood was to solve the following maximin optimisation problem:
\begin{equation*}
\max_{\theta} \min_{\alpha \in \mathbb{R}^N} -\frac{1}{2}\log \det (\hat{K}+\sigma^2 I) + \frac{1}{2}\alpha^\top(K+\sigma^2 I)\alpha-\alpha^\top y
\end{equation*}
One way to solve this cheaply would be by coordinate descent, where one maximises with respect to $\theta$ fixing $\alpha$, then minimises with respect to $\alpha$ fixing $\theta$. However $\sigma$ tends to blow up in practice. This is because the expression is $O(-\log \sigma^2 + \sigma^2)$ for fixed $\alpha$, hence maximising with respect to $\sigma$ pushes it towards infinity.

An alternative is to sum the two upper bounds above to get the upper bound
\begin{equation*}
-\frac{1}{2}\log \det (\hat{K}+\sigma^2 I) -\frac{1}{2}y^\top(K+(\sigma^2 +\Tr(K-\hat{K}))I)^{-1}y
\end{equation*}
However we found that maximising this bound gives quite a loose upper bound unless $m=O(N)$. Hence this upper bound is not very useful.

\section{Random Fourier Features}
\label{apd:rff}
Random Fourier Features (RFF) (a.k.a. Random Kitchen Sinks) was introduced by \cite{rahimi2007random} as a low rank approximation to the kernel matrix. It uses the following theorem
\begin{theorem}[Bochner's Theorem \cite{rudin1964fourier}]\label{thm:bochner}
A stationary kernel k(d) is positive definite if and only if k(d) is the Fourier transform of a non-negative measure.
\end{theorem}
to give an unbiased low-rank approximation to the Gram matrix $K=\mathbb{E}[\Phi^\top\Phi]$ with $\Phi \in \mathbb{R}^{m \times N}$. A bigger $m$ lowers the variance of the estimate. Using this approximation, one can compute determinants and inverses in $O(Nm^2)$ time. In the context of kernel composition in \ref{sec:abcd}, RFFs have the nice property that samples from the spectral density of the sum or product of kernels can easily be obtained as sums or mixtures of samples of the individual kernels (see Appendix \ref{apd:rf}). We use this later to give a memory-efficient upper bound on the exact log marginal likelihood in Appendix \ref{apd:rff_ub}.

\section{Random Features for Sums and Products of Kernels}\label{apd:rf}
For RFF the kernel can be approximated by the inner product of random features given by samples from its spectral density, in a Monte Carlo approximation, as follows:
\begin{align*}
k(x-y) &= \int_{\mathbb{R}^D} e^{iv^\top (x-y)} d\mathbb{P}(v) \\
&\propto \int_{\mathbb{R}^D} p(v)e^{iv^\top (x-y)} dv \\
&= \mathbb{E}_{p(v)}[e^{iv^\top x}(e^{iv^\top y})^*] \\
&= \mathbb{E}_{p(v)}[Re(e^{iv^\top x}(e^{iv^\top y})^*)] \\
& \approx \frac{1}{m} \sum_{k=1}^m Re(e^{i{v_k}^\top x}(e^{i{v_k}^\top y})^*) \\
& = \mathbb{E}_{b,v} [\phi(x)^\top  \phi(y)]
\end{align*}
where $\phi(x) = \sqrt{\frac{2}{m}}(cos({v_1}^\top x+b_1),\ldots,cos({v_m}^\top x+b_m))$ with spectral frequencies $v_k$ iid samples from $p(v)$ and $b_k$ iid samples from $U[0,2\pi]$. \\ 
Let $k_1, k_2$ be two stationary kernels, with respective spectral densities $p_1,p_2$ so that \\
$k_1(d)=a_1\hat{p_1}(d), k_2(d)=a_2\hat{p_2}(d)$, where $\hat{p}(d):=\int_{\mathbb{R}^D} p(v)e^{iv^\top d}dv$. We use this convention as the Fourier transform. Note $a_i=k_i(0)$.
\begin{align*}
    (k_1+k_2)(d)&=a_1\int p_1(v)e^{iv^\top d} dv + a_2\int p_2(v)e^{iv^\top d} dv\\ &=(a_1+a_2)\hat{p_+}(d)
\end{align*}
where $p_+(v)=\frac{a_1}{a_1+a_2}p_1(v)+\frac{a_2}{a_1+a_2}p_2(v)$, a mixture of $p_1$ and $p_2$. So to generate RFF for $k_1+k_2$, generate $v \sim p_+$ by generating 
$v \sim p_1 \text{ w.p. } \frac{a_1}{a_1+a_2} \text{ and } v \sim p_2 \text{ w.p. } \frac{a_2}{a_1+a_2}$ \\
Now for the product, suppose
\[(k_1\cdot k_2)(d)=a_1a_2\hat{p_1}(d)\hat{p_2}(d)=a_1a_2\hat{p_*}(d)\]
Then $p_*(d)$ is the inverse fourier transform of $\hat{p_1}\hat{p_2}$, which is the convolution $p_1*p_2(d):=\int_{\mathbb{R}^D} p_1(z)p_2(d-z)dz$. So to generate RFF for $k_1\cdot k_2$, generate $v \sim p_*$ by generating $v_1 \sim p_1, v_2 \sim p_2$ and setting $v=v_1+v_2$. \\
This is not applicable for non-stationary kernels, such as the linear kernel. We deal with this problem as follows:

Suppose $\phi_1,\phi_2$ are random features such that \\ $k_1(x,x')=\phi_1(x)^\top \phi_1(x'),\phi_2(x)^\top \phi_2(x'), \phi_i:\mathbb{R}^D \rightarrow \mathbb{R}^m$.\\
It is straightforward to verify that
\begin{align*}
    (k_1+k_2)(x,x')&=\phi_+(x)^\top \phi_+(x')  \\
    (k_1\cdot k_2)(x,x')&=\phi_*(x)^\top \phi_*(x')
\end{align*}
where $\phi_+(\cdot)=(\phi_1(\cdot)^\top ,\phi_2(\cdot)^\top )^\top$ and $\phi_*(\cdot)=\phi_1(\cdot) \otimes \phi_2(\cdot)$. However we do not want the number of features to grow as we add or multiply kernels, since it will grow exponentially. We want to keep it to be $m$ features. So we subsample $m$ entries from $\phi_+$ (or $\phi_*$) and scale by factor $\sqrt{2}$ ($\sqrt{m}$ for $\phi_*$), which will still give us unbiased estimates of the kernel provided that each term of the inner product $\phi_+(x)^\top \phi_+(x')$ (or $\phi_*(x)^\top \phi_*(x')$) is an unbiased estimate of $(k_1+k_2)(x,x')$(or $(k_1 \cdot k_2)(x,x')$).  \\
%So it is convenient to use first form of RFF. With second form, need grouping
This is how we generate random features for linear kernels combined with other stationary kernels, using the features $\phi(x)=\frac{\sigma}{\sqrt{m}}(x, \ldots , x)^\top $.

\section{Spectral Density for PER}\label{apd:rqper}
From \cite{solin2014explicit}, we have that the spectral density of the PER kernel is:
\begin{align*}
    \sum_{n=-\infty}^{\infty} \frac{I_n(l^{-2})}{\exp(l^{-2})} \delta \bigg(v-\frac{2\pi n}{p}\bigg)
\end{align*}
where  $I$ is the modified Bessel function of the first kind.

\section{An upper bound to NLD using Random Fourier Features} \label{apd:rff_ub}
Note that the function $f(X)=-\log \det(X)$ is convex on the set of positive definite matrices \cite{boyd2004convex}. Hence by Jensen's inequality we have, for $\Phi^\top\Phi$ an unbiased estimate of $K$:
\begin{align*}
-\frac{1}{2}\log\det (K+\sigma^2 I) &=f(K+\sigma^2 I) \\
&=f(\mathbb{E}[\Phi^\top\Phi+\sigma^2 I]) \\
&\leq \mathbb{E}[f(\Phi^\top\Phi+\sigma^2 I)]
\end{align*}
Hence $-\frac{1}{2}\log \det (\Phi^\top\Phi+\sigma^2 I)$ is a stochastic upper bound to NLD that can be calculated in $O(Nm^2)$. An example of such an unbiased estimator $\Phi$ is given by RFF. 

\section{Further Plots} \label{apd:plots}
\begin{figure*}[h!]
  \centering
  \subfloat[Mauna: fix inducing points]{\includegraphics[width=0.5\textwidth]{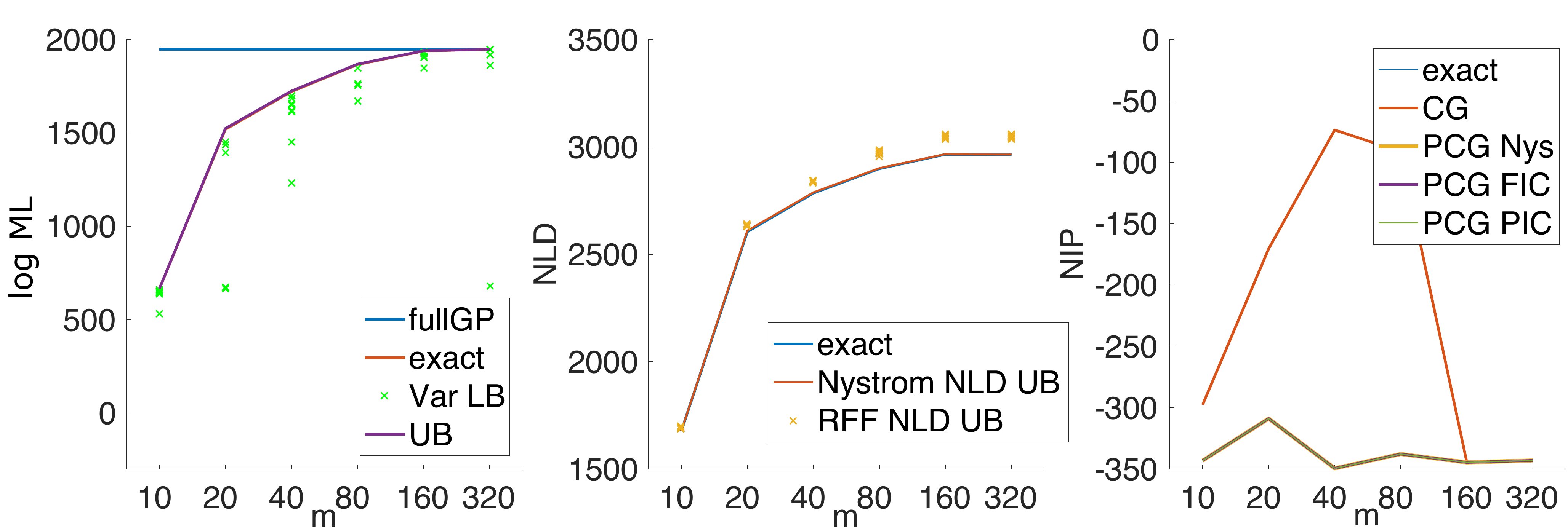}\label{fig:exp_mauna_subset_half}}
  \hfill
  \subfloat[Mauna: learn inducing points]{\includegraphics[width=0.5\textwidth]{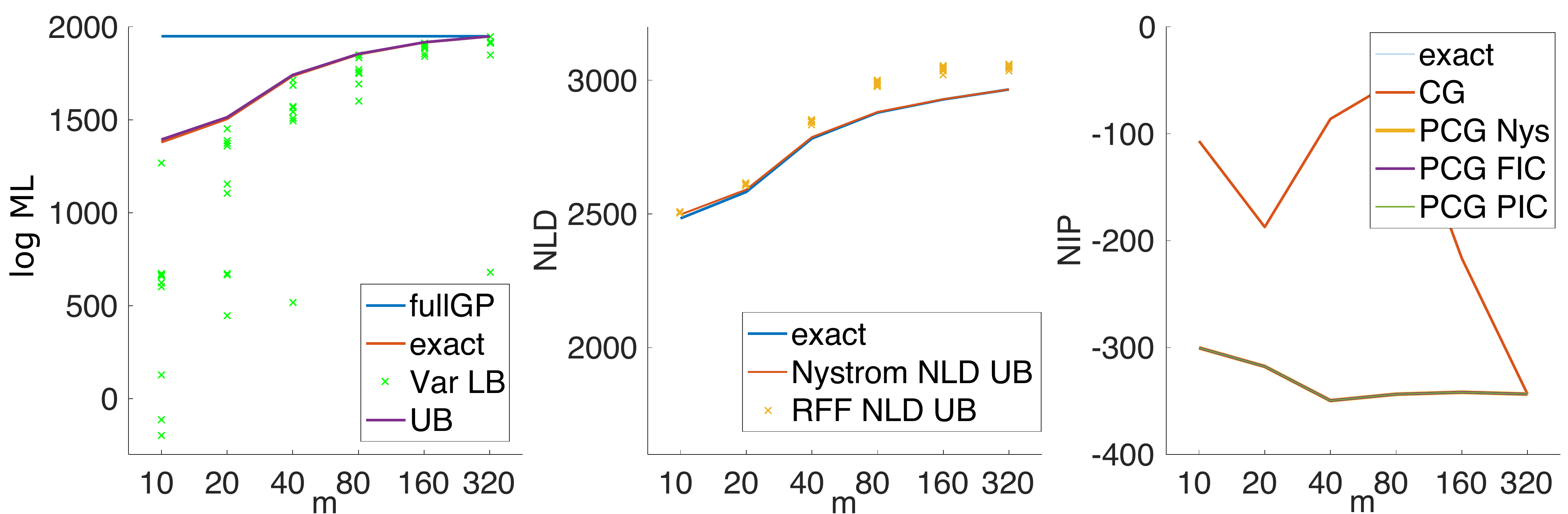}\label{fig:exp_mauna_learn_ind_pts}}
  \caption{Same as \ref{fig:exp_solar_subset_half} and \ref{fig:exp_solar_learn_ind_pts} but for Mauna Loa data.}
\end{figure*}

\begin{figure*}[h!]
  \centering
  \subfloat[Concrete: fix inducing points]{\includegraphics[width=0.5\textwidth]{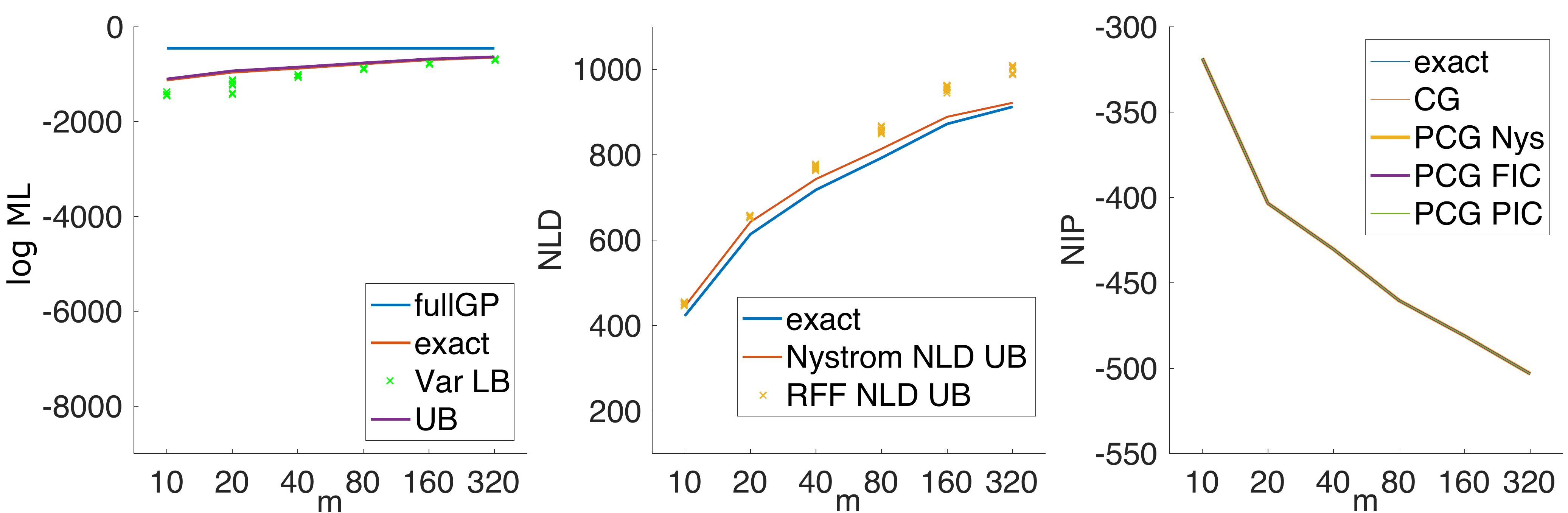}\label{fig:exp_concrete_subset_half}}
  \hfill
  \subfloat[Concrete: learn inducing points]{\includegraphics[width=0.5\textwidth]{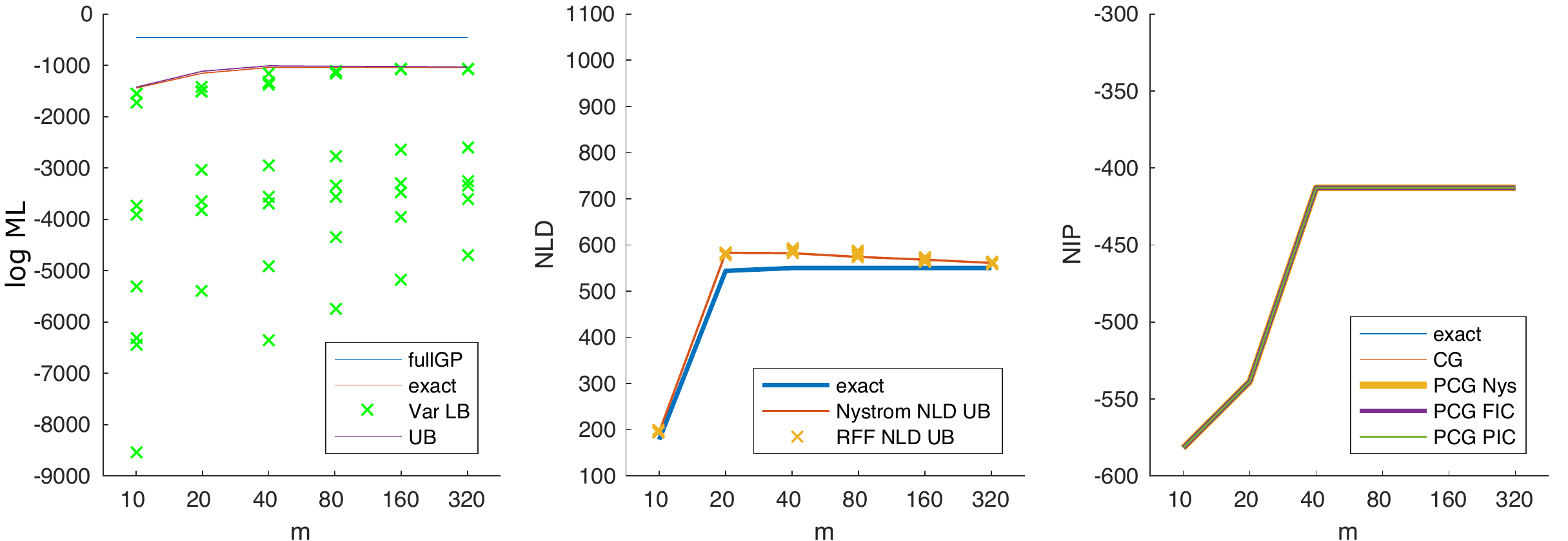}\label{fig:exp_concrete_learn_ind_pts}}
  \caption{Same as \ref{fig:exp_solar_subset_half} and \ref{fig:exp_solar_learn_ind_pts} but for Concrete data.}
\end{figure*}

\begin{figure*}[ht!]
  \centering
  \includegraphics[width=\linewidth]{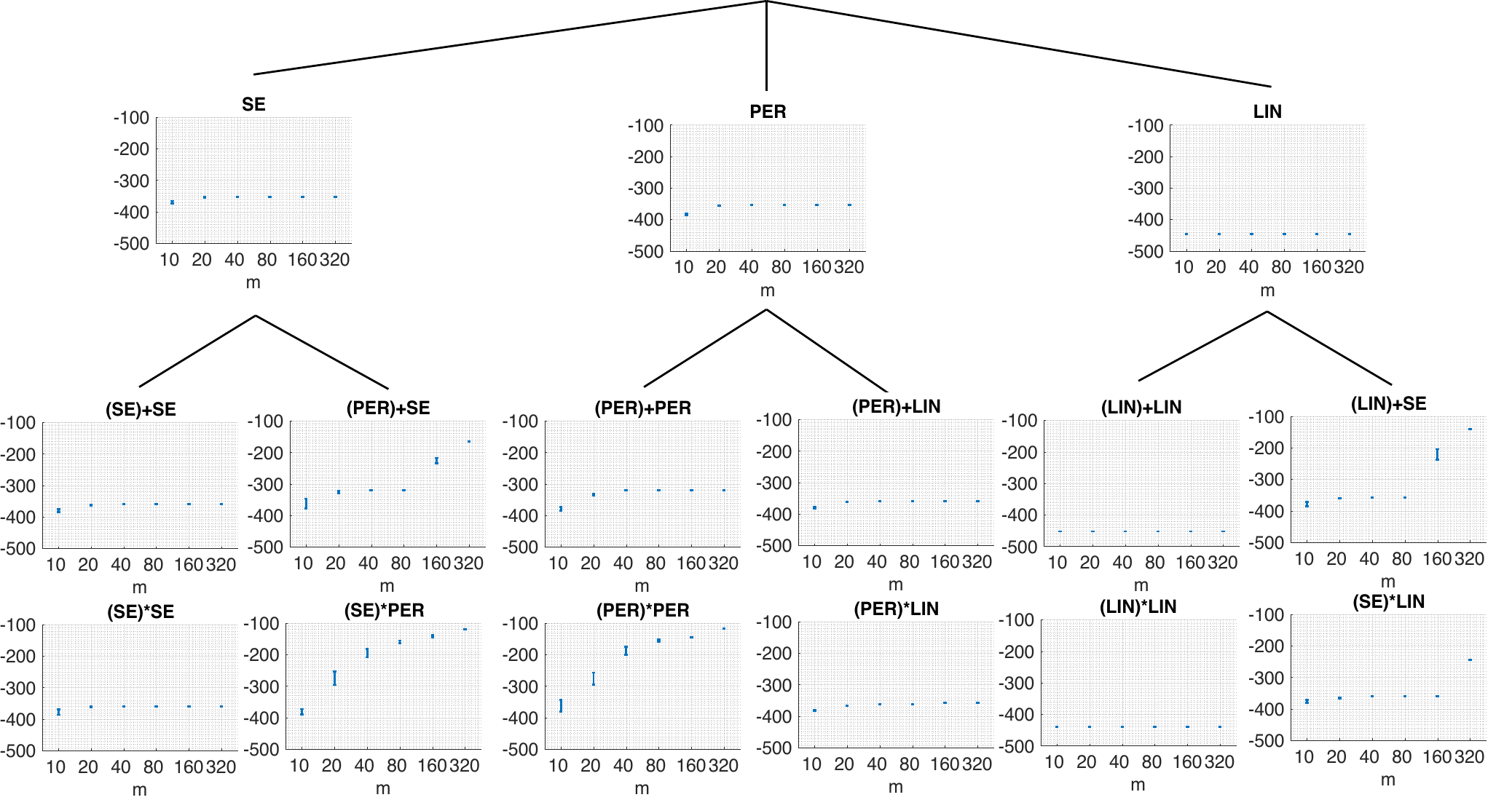}
  \caption{Kernel search tree for SKC on solar data up to depth 2. We show the upper and lower bounds for different numbers of inducing points $m$.}\label{fig:solar_tree}
\vspace*{-3mm}
\end{figure*}

\begin{figure*}[ht!]
  \centering
  \includegraphics[width=\linewidth]{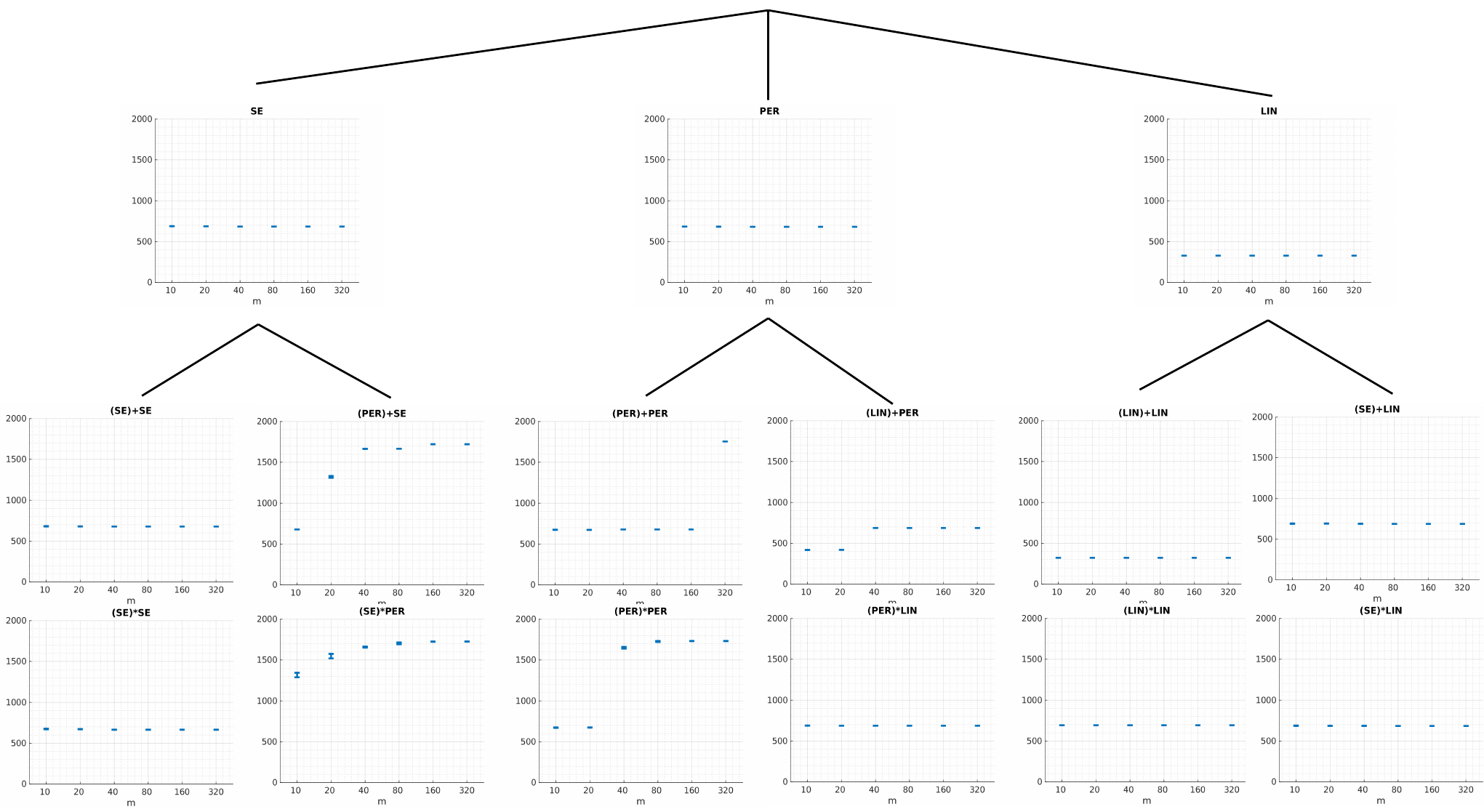}
  \caption{Same as Figure \ref{fig:solar_tree} but for Mauna data.}\label{fig:mauna_tree}
\end{figure*}

\begin{figure*}[ht!]
  \centering
  \includegraphics[width=\linewidth]{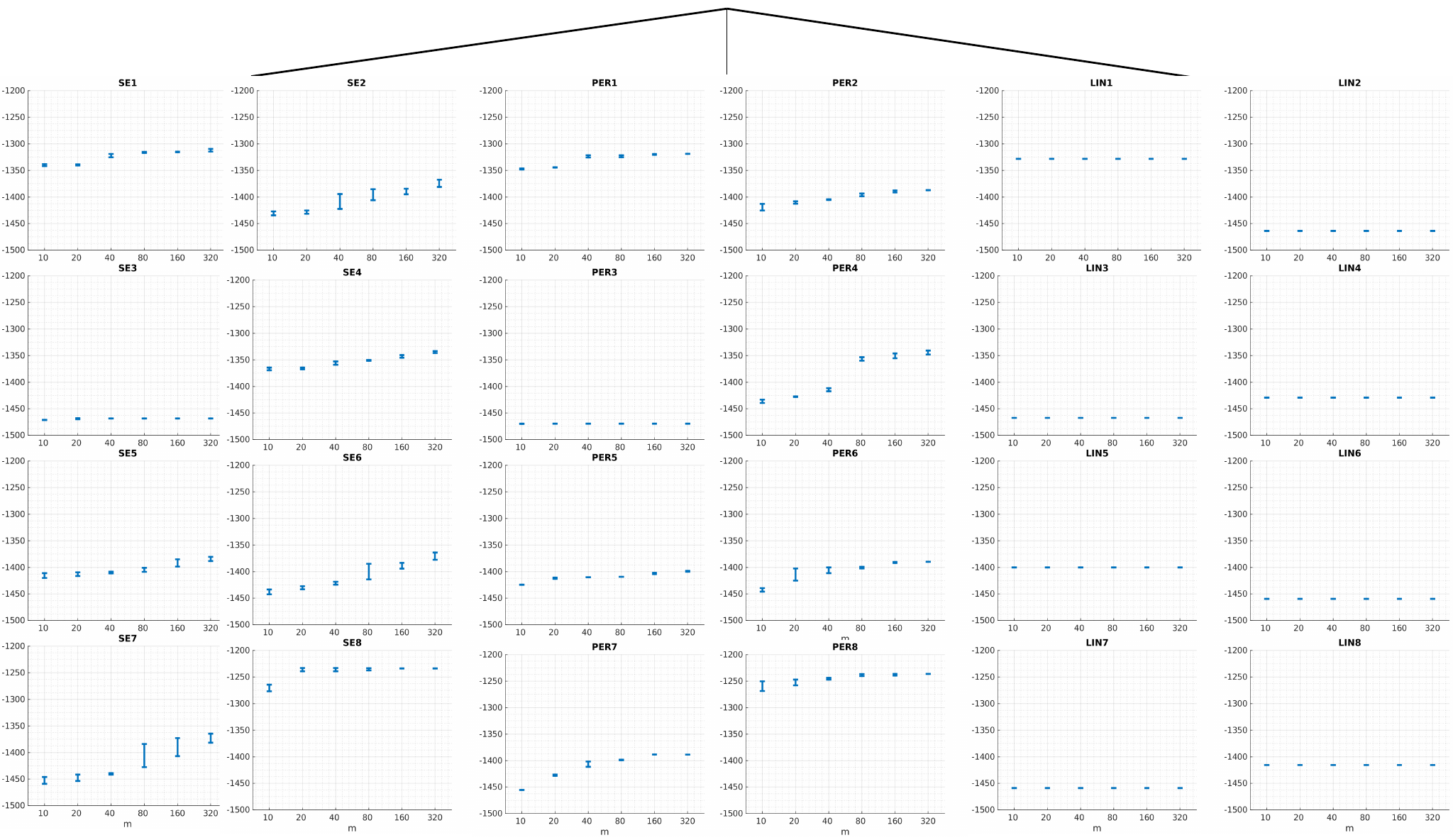}
  \caption{Same as Figure \ref{fig:solar_tree} but for concrete data and up to depth 1.}\label{fig:concrete_tree}
\end{figure*}

\end{document}